\pdfoutput=1

\documentclass[11pt]{article}
\usepackage{hhline}

\usepackage[final]{coling}

\usepackage{times}
\usepackage{latexsym}
\usepackage{tikz}
\usepackage{graphicx}  
\usepackage{caption}
\usepackage{array}

\newcolumntype{P}[1]{>{\centering\arraybackslash}p{#1}}

\newcommand*\circled[1]{\tikz[baseline=(char.base)]{
            \node[shape=circle,draw,inner sep=1pt] (char) {#1};}}
            
\usepackage[T1]{fontenc}

\usepackage[utf8]{inputenc}

\usepackage{microtype}
\usepackage{booktabs}

\usepackage{inconsolata}

\usepackage{graphicx}
\usepackage{svg}

\title{
RISCORE: Enhancing In-Context Riddle Solving in Language Models through Context-Reconstructed Example Augmentation 
}

\author{
    Ioannis Panagiotopoulos,  Giorgos Filandrianos, Maria Lymperaiou, Giorgos Stamou\\ 
    School of Electrical and Computer Engineering,  AILS Laboratory\\
    National Technical University of Athens \\
    \texttt{\href{mailto:yiannispn@gmail.com}{yiannispn@gmail.com}, \{\href{mailto:geofila@islab.ntua.gr}{geofila}, \href{mailto:marialymp@islab.ntua.gr}{marialymp}\}@islab.ntua.gr, \href{mailto:gstam@cs.ntua.gr}{gstam@cs.ntua.gr}}\\
}

\begin{document}
\maketitle
\begin{abstract}

Riddle-solving requires advanced reasoning skills, pushing Large Language Models (LLMs) to engage in abstract thinking and creative problem-solving, often revealing limitations in their cognitive abilities. In this paper, we examine the riddle-solving capabilities of LLMs using a multiple-choice format, exploring how different prompting techniques impact performance on riddles that demand diverse reasoning skills. To enhance results, we introduce RISCORE (\textbf{RI}ddle \textbf{S}olving with \textbf{CO}ntext \textbf{RE}contruciton), a novel fully automated prompting method that generates and utilizes contextually reconstructed sentence-based puzzles in conjunction with the original examples to create few-shot exemplars. Our experiments demonstrate that RISCORE significantly improves the performance of language models in both vertical and lateral thinking tasks, surpassing traditional exemplar selection strategies across a variety of few-shot settings \footnote{Code for
our experiments is available at: \url{https://github.com/GiannisPana/RISCORE}}.

\end{abstract}

\section{Introduction}
Reasoning in Natural Language Processing has been a field of increasing interest, especially since the surge of Large Language Models (LLMs), which showcase several reasoning shortcomings \cite{nlp-reasoning, qiao-etal-2023-reasoning}. The logical gap between humans and LLMs' advanced reasoning is further exposed when probing the puzzle solving capabilities of such models \cite{giadikiaroglou2024puzzlesolvingusingreasoning}, denoting the necessity of intricate datasets \cite{lin-etal-2021-riddlesense, szomiu2021puzzle, zhang2022birdqabilingualdatasetquestion} that test and enhance complex model reasoning. In terms of advanced reasoning skills, lateral thinking datasets such as BrainTeaser \cite{jiang-etal-2023-brainteaser} stress the creative reasoning capabilities of models, requiring them to defy obvious logical associations and approach a more abstract way of thinking.

\begin{figure}[h!]
    \centering
    \includegraphics[width=\linewidth]{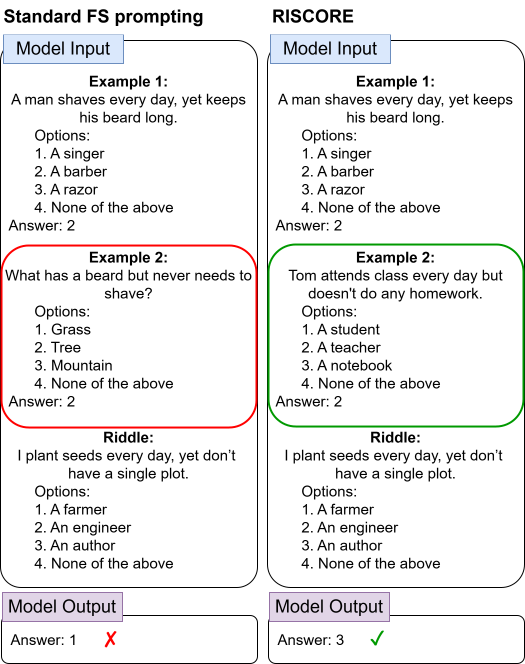}
    \caption{Standard \textbf{FS} prompting vs RISCORE prompting: by encouraging reasoning-based selection of exemplars (riddle in green) in place of semantic similarity selection (riddle in red), the model unlocks the reasoning pattern, guided towards the correct answer.}
    \label{fig:teaser}
\end{figure}

At the same time, prompting is a widespread technique of extracting reasoning patterns from LLMs \cite{qiao-etal-2023-reasoning}. Especially Chain-of-Thought (CoT) prompting \cite{wei2023chainofthoughtpromptingelicitsreasoning}, as naturally designed to trigger the extraction of reasoning chains, is indeed proven to be one successful and viable prompting strategy in practice, even though its functionality is only applicable on larger models. Other promising techniques, such as in-context learning, are highly capable when the right data samples are utilized as exemplars, but fail to properly drive LLMs otherwise \cite{dong2024surveyincontextlearning}. While similarity-driven exemplar selection serves as a well-established rule of thumb \cite{liu-etal-2022-makes, qin2024incontextlearningiterativedemonstration, dong2024surveyincontextlearning}, the notion of similarity becomes more obscure when concerning reasoning patterns: for example, the sentences \textit{``A man shaves everyday, yet keeps his beard long''} and \textit{``Tom attends class every day but doesn’t do any homework''} \cite{jiang-etal-2023-brainteaser} in fact represent the exact same reasoning pattern, despite being rather distant semantically.

\begin{figure}[h!]
    \centering
    \includegraphics[width=\linewidth]{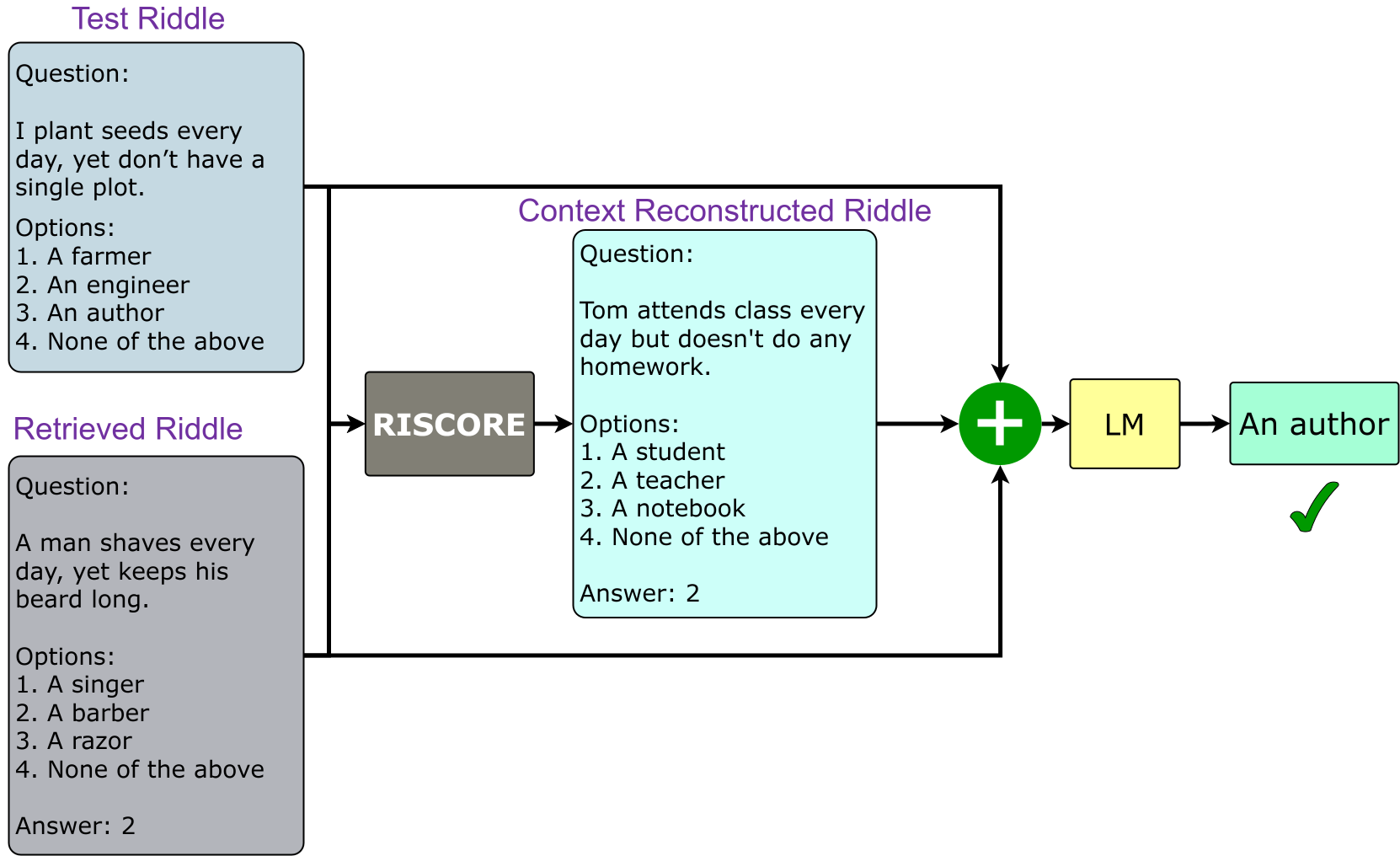}
    \caption{An overview of RISCORE, where the reconstructed instances, along with their original counterparts, are incorporated as exemplars in the few-shot setting to enhance the model's riddle solving ability.}
    \label{fig:method}
\end{figure}

Driven by this observation, we assume that by ``stripping'' a sentence from linguistic features, while focusing on pure reasoning patterns prevalent in sentences, we can provide the model with the critical information to unlock the desired underlying reasoning pattern. Sentences that represent the same reasoning pattern are provided as ``context reconstruction'' counterparts in BrainTeaser data \cite{jiang-etal-2023-brainteaser} or can be automatically generated using our proposed method. Therefore, incorporating such reconstructed instances as few-shot exemplars, together with their original counterparts, offers an alternative few-shot strategy that prioritizes reasoning similarity and analogy.
An example of the method is provided in Figure \ref{fig:method}. To this end, we contribute to the following:


\begin{itemize}

    \item We experimentally verify that providing a riddle along with its context reconstruction can enhance performance on both lateral and vertical thinking problems.
    

    \item We propose RISCORE (Example provided in Figure \ref{fig:teaser}), a novel prompting method designed to enhance the in-context riddle-solving capabilities of LLMs. As a supplement to this, we introduce an algorithm for generating contextually reconstructed multiple-choice riddles.

    \item We compare RISCORE against a wide array of popular prompting techniques, highlighting its effectiveness across several prompting alternatives and models of varying sizes.
    
\end{itemize}

\section{Related work}


\paragraph{Reasoning with Language Models}
Multiple reasoning senses have been studied over the years using language models, including commonsense \cite{sap-etal-2020-commonsense}, arithmetic \cite{luo2024improvemathematicalreasoninglanguage}, abductive \cite{zhao-etal-2023-abductive}, inductive \cite{HAN2024101155}, deductive \cite{sanyal-etal-2022-fairr}, analogical reasoning \cite{sultan-shahaf-2022-life} and others. Linear thinking processes that exploit rules and logic, termed as ``vertical thinking'', have been previously explored within popular datasets \cite{lin-etal-2021-riddlesense, piqa}, unveiling interesting reasoning patterns of language models.
On the other hand, creative thinking has been widely underexplored and often purposefully excluded from reasoning benchmarks \cite{conceptnet, atomic}, leading to a significant gap in literature, especially given the emergent capabilities that larger models present \cite{wei2022emergentabilitieslargelanguage}. Puzzle solving closely lies to creative reasoning \cite{giadikiaroglou2024puzzlesolvingusingreasoning, lin-etal-2021-riddlesense, zhang2022birdqabilingualdatasetquestion}, since out-of-the-box thinking is mainly required. Going one step further, overriding default presuppositions and associations occurring when reasoning leads to ``lateral thinking'' processes, leading to more tricky puzzles, as demonstrated in the BrainTeaser dataset \cite{jiang-etal-2023-brainteaser} for the first time. In our work, we focus on probing vertical and lateral puzzle-solving reasoning abilities of language models via prompting.

\begin{figure*}[ht!]
    \centering
    \includegraphics[width=\textwidth]{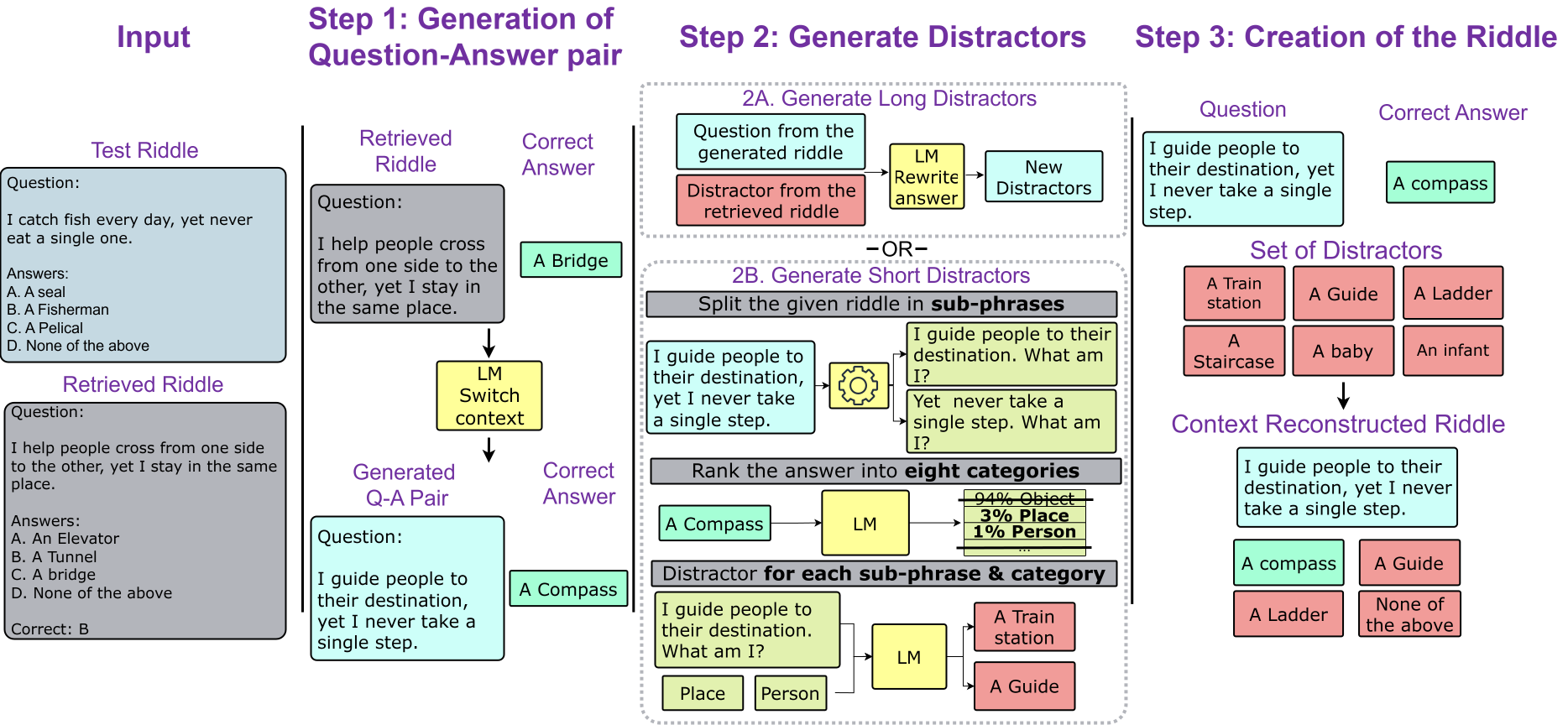}
    \caption{An overview of the automated method for generating a context-reconstructed riddle.}
    \label{fig:automated-gen-method}
\end{figure*}

\paragraph{LLMs and prompting} Discovering reasoning patterns in LLMs is often performed via various types of prompting
\cite{qiao-etal-2023-reasoning}. Zero-shot strategies harness simple but intuitive instructions, such as ``Let's think step-by-step'' that successfully improve LLM reasoning \cite{kojima2023largelanguagemodelszeroshot}; however, the search space of such magic prompts seems indefinite. Querying for intermediate reasoning steps was framed in the seminal Chain-of-Thought (CoT) prompting \cite{wei2023chainofthoughtpromptingelicitsreasoning}, proving that it is sufficient to elicit reasoning in larger models. Nevertheless, the placement of demonstrations in few-shot prompting faces several challenges \cite{dong2024surveyincontextlearning}; specifically, the way of selecting these exemplars themselves is prone to instabilities \cite{lu-etal-2022-fantastically, min-etal-2022-rethinking}, proving the non-negotiable need for developing optimal exemplar selection techniques. Similarity-based retrieval is established as a default selection technique \cite{liu-etal-2022-makes, qin2024incontextlearningiterativedemonstration, dong2024surveyincontextlearning}, with related enhancement optimizing the ordering of selected exemplars \cite{wu-etal-2023-self}. With a focus on reasoning tasks, complexity of reasoning paths \cite{fu2023complexitybasedpromptingmultistepreasoning} or diversity of exemplars \cite{zhang2022automaticchainthoughtprompting} have already proven their merits, shifting the weight toward uncovering the patterns hidden in the data themselves that are able to advance reasoning, rather than similarity-driven placement based on semantics. Inspired by the above, we craft exemplars that promote the hidden reasoning patterns rather than the semantics of related data samples, maintaining simple similarity-based retrieval for their placement; by outperforming other prompting techniques, we demonstrate that highlighting those reasoning patterns is adequate for advanced LLM reasoning, without further few-shot engineering.

\section{Method}
\label{sec:method}






\subsection{RISCORE prompting}
Consider the following two riddles: \textit{R1: ``A man shaves every day, yet keeps his beard long''} and \textit{R2: ``What has a beard but never needs to shave?''}. Although these riddles are semantically similar, follow the same structure, and refer to the same objects, their reasoning processes differ. In \textit{R1}, where the answer is ``A barber,'' the word ``beard'' is used in the context of human grooming and personal appearance.
In \textit{R2}, where the answer is ``A tree,'' ``beard'' is used  in the botanical or natural context, referring to the ``beard'' of certain trees, such as the ``oak''. Thus, when using these riddles as context for a new riddle—such as \textit{``I plant seeds every day, yet don't have a single plot''}—the answer will depend on the interpretation of the phrase ``plant seeds'', based on its contextual framing.
In a creative context, the correct answer is "An author". Authors ``plant seeds'' of ideas through their writing but do not have a literal ``plot'', unless it refers to the abstract concept of a story plot, which adds layers of interpretation. On the other hand, ``A farmer'' is not the correct answer because a farmer typically works with physical plots of land to plant seeds, which contradicts the phrase ``yet don’t have a single plot.''

In this example, however, instead of using \textit{R2}, a contextually reconstructed version of \textit{R1}—\textit{R3: ``Tom attends class every day but doesn’t do any homework''}—would provide a clearer reasoning process for the model. Although \textit{R2} is semantically closer to \textit{R1}, its use of ``beard'' in a natural or botanical context introduces a different reasoning pathway, illustrating how the contextual framing of a riddle can clarify or obscure its intended answer.


Building on the previous example, we propose the RISCORE prompting method, designed to enhance the in-context riddle-solving abilities of LLMs. The RISCORE method supplements each exemplar in \textbf{FS} learning with a contextually reconstructed version of itself. This preserves the desired reasoning process while only altering its context. This approach enables the model to delineate a clear and coherent reasoning trajectory, which it can then follow to effectively solve new riddles.

As illustrated in Figure \ref{fig:method}, our approach extends existing few-shot (FS) methods \cite{dong2024surveyincontextlearning, wang-etal-2024-learning, sultan-shahaf-2022-life}, with RISCORE not involved in the exemplar selection process\footnote{We utilize semantic similarity for optimal exemplar selection, but other methods from literature may be used.}. Specifically, the goal of our method is to \textbf{augment FS samples} with automatically generated context-reconstructed examples (detailed in Section \ref{sec:method_aug}). The inclusion of these context-adapted examples has, in most cases, demonstrated greater impact in improving model performance, even outperforming the use of real examples extracted directly from the dataset (see Section \ref{sec:full_results}).

\subsection{Method for generating context reconstructions}
\label{sec:method_aug}
In this section, we present our approach for generating high-quality contextually reconstructed riddles to serve as few-shot exemplars for a model in a Multiple-Choice Question Answering (MQA) format. These exemplars, combined with the original pairs, aim to enhance the model's performance on lateral and vertical thinking tasks.
Our method builds on the semi-automated pipeline proposed in the BrainTeaser paper \cite{jiang-etal-2023-brainteaser}, but we advance this framework by fully harnessing the capabilities of LLMs to automate the process. An overview of the automated method for generating a context-reconstructed riddle is  provided in Figure \ref{fig:automated-gen-method}. 

\paragraph{Step 1: Generation of Question-Answer pair} 
The first step involves generating one contextually reconstructed Question-Answer pair per selected instance, temporarily ignoring distractors. \textit{
Distractors} constitute the incorrect answer options presented alongside the correct answer in multiple-choice settings, designed to appear plausible and test the depth of understanding or reasoning of the model.
To achieve this, we provide the LLM with the riddle, the correct answer, and a system prompt that outlines the task. The prompt instructs the model to analyze the given Question-Answer pair, understand the riddle, identify the reasoning process that links the question to the answer, and then generate a similar riddle that follows the same reasoning process, along with a corresponding correct answer. Since the model is provided with both the riddle and the correct answer, understanding the riddle becomes easier than solving it independently \cite{sultan-shahaf-2022-life}. To further refine this approach, we apply the method in both Zero-shot (\textbf{ZS}) and Few-shot (\textbf{FS}) settings, using pre-existing pairs of original and contextually reconstructed questions from BrainTeaser. The prompts used in this process can be found in App. \ref{sec:qa_construction}.
After generating the Question-Answer pair, a filtering procedure was conducted to ensure the quality of the instance and its alignment with the riddles in the dataset. This process involves applying rules concerning both the structure of the riddle and the answer. These rules are dataset-specific and can be adjusted to suit different datasets, ensuring adaptability across various contexts. Further analysis regarding this step can be found in App. \ref{sec:qa_construction_det}.

\paragraph{Step 2: Generation of the distractors} 

This process involves generating incorrect answers to complement the multiple-choice options. While this task may initially appear straightforward, it presents several significant challenges. First, the distractors must not only match the number of original options, but also be guaranteed to be incorrect when compared to the correct answer. Additionally, they should not deviate excessively from the correct answer, as significant divergence could undermine the validity and challenge of the riddle. It is also essential to account for cases where the correct answer is ``None of the above'', as observed in the BrainTeaser dataset, ensuring that the other three alternatives are wrong enough. 

The length of the answers also plays a crucial role in this process. For instance, BrainTeaser includes mostly answers with more than four or five words, while RiddleSense provides typically single-word answers. These variations in length introduce additional challenges in crafting effective distractors. To address this, we propose two distinct methods for generating distractors: one for long and another for short distractors, respectively.

\paragraph{Generation of Long Distractors}
In this approach, the original distractors of each riddle are given to an LLM alongside the new Question generated in Step 1. For each distractor (processed one at a time, except for ``None of the above''), the model is prompted to modify its context by integrating it with the context of the new riddle. Regardless of the quality of this integration, the distractor is guaranteed to be incorrect, though lower-quality incorporation may slightly reduce its overall effectiveness. 
If the original riddle's answer differs from "None of the above", but the new riddle's answer matches it, an additional distractor must be generated. In such cases, we simply prompt the LLM to generate a new distractor based on the generated Question-Answer pair. Details of the aforementioned step can be found in App. \ref{sec:dist_bt}.

\paragraph{Generation of Short Distractors}
The second method for generating distractors is applied to datasets like RiddleSense, which contain single-word answers. In this scenario, modifying a single word like ``water'' to fit the riddle’s context presents a challenge. To address this, we first classify the generated Answer of Step 1 into a set of mutually exclusive categories (details in App. \ref{sec:dist_rs}), which informs the generation of distractors. For instance, in the riddle ``I have a beard, but I never shave. What am I?'' (where the correct answer is ``A tree''), the system would classify the answer under the category ``Nature''.

Next, to ensure the distractor is incorrect, we break the riddle into smaller phrases. We also detect the presence of interrogative words, and if the question is purely descriptive without a direct query, we append the phrase ``What am I?'' at the end. For the previous example, we split the riddle into two parts: ``I have a beard, what am I?'' and ``I never shave. What am I?'' These phrases are then provided as inputs to the LLM, and only answers that differ from the correct one are retained. Since the LLM has access only to a phrase, we expect the generated answer to be incorrect. However, this is not always the case. In the example above, the response to ``I have a beard, what am I?'' might be ``An oak,'' which is also a valid answer for the riddle.

To address this issue, we leverage the classification of the Answer to guide distractor generation. Specifically, we provide the LLM with two closely related categories, excluding the correct answer’s category (e.g., ``person'' and ``place''). The model is then prompted, through a system-user instruction, to generate a valid answer within the given context and category. This approach ensures that the generated distractors are both incorrect and contextually relevant, thereby preserving the difficulty of the riddle.
After applying a filtering stage to ensure the distractors are distinct, we use WordNet \cite{miller-1992-wordnet} to augment them if their quantity is insufficient.
Further details can be found in App.  \ref{sec:dist_rs}.
Additionally, for both methods, the same filtering procedure applied to the correct answer is also used to ensure the quality of the distractors.

\paragraph{Step 3: Creation of the Riddle}
In the final step of this pipeline, we use the Question-Answer pair, randomly shuffle the distractor set, select the required number of distractors, and place the correct answer in a randomly assigned position, so that a possible position bias is eliminated. More details regarding the Prompting process for our method can be found in App. \ref{sec:pairs_prompting}.

\section{Experiments}

We evaluate various LLMs using different datasets with our method and compare their performance against multiple baselines. 

\subsection{Datasets}
For our experiments, we utilize two different datasets, namely BrainTeaser and RiddleSense, addressing vertical and lateral reasoning respectively. A comprehensive analysis of the datasets used in our experiments can be found in App \ref{sec:dataset}.

We select BrainTeaser primarily for two key reasons. First, BrainTeaser provides high-quality, manually crafted context reconstructions, which establish an upper bound on the model's performance when these are incorporated into the input. This upper bound also serves as a benchmark for evaluating the quality of riddles generated by the automated method. Second, the dataset consists of riddles that test the lateral thinking abilities of the models, a reasoning process that has proven as a challenging cognitive process for LLMs \cite{giadikiaroglou2024puzzlesolvingusingreasoning}.
Additionally, experiments using RiddleSense are conducted, emphasizing on the vertical thinking abilities of models, while verifying our method's broad applicability across different types of riddles. 
Given that hand-crafted context reconstructions are not available for RiddleSense, the automated method was utilized without ground-truth reconstructed instances in this set of experiments.


\subsection{Baselines}

In our study, we use various prompting techniques as baselines, including zero-shot (\textbf{ZS}), few-shot (\textbf{FS}), and Chain of Thought (\textbf{CoT}) methods.
A detailed analysis of each technique is provided in App. \ref{sec:prompts}.

\paragraph{Chain-of-thought (\textbf{CoT ZS})} In this experiment, we prompt the model to answer the multiple-choice riddle in a step-by-step manner, following the method proposed in \citet{wei2023chainofthoughtpromptingelicitsreasoning}, without demonstrating any examples. This approach allows us to assess the models' riddle-solving abilities in a zero-shot setting.

\paragraph{Few-Shot (\textbf{FS})} To assess the influence of example-based learning, we incorporate \textbf{FS} prompting with varying numbers of examples: 2, 4, and 8 shots. We limit our tests to 8 shots since it has been proven that reasoning abilities of LLMs  are deteriorating with longer inputs \cite{levy2024tasktokensimpactinput}. Two distinct strategies were used for selecting these examples: random selection, referred to as \textbf{Rand}, and selecting riddles with minimal semantic distance 
from the test riddle using \citet{zhang2024mgtegeneralizedlongcontexttext}\footnote{\url{https://huggingface.co/Alibaba-NLP/gte-large-en-v1.5}} as the semantic similarity model, referred to as \textbf{Sim}.


\paragraph{Few-shot with CoT Explanations (\textbf{CoT FS})} A straightforward way to aid the model in understanding an exemplar is by providing a supplementary explanation. In this experiment, we explore this approach by offering explanations alongside the few-shot exemplars. This allows us to evaluate whether these crafted explanations can improve the models' ability to comprehend and follow the required reasoning process. The explanations are generated using a semi-automated process. First, for each exemplar in the \textbf{FS} setting, we use ChatGPT\footnote{Specifically, we used version gpt-3.5-turbo-0125.} to generate an explanation for the riddle based on the correct answer, following the method proposed in \citet{wei2023chainofthoughtpromptingelicitsreasoning}. These explanations were then manually curated to ensure alignment with human interpretations of the reasoning process required for the riddle. Due to the labor-intensive process of annotating the explanations, this experiment is conducted using only the 2, 4, and 8-shot settings with a randomly selected sampling strategy.

\subsection{RISCORE}

RISCORE is applied as follows. For each example used in the \textbf{FS} setting, the corresponding contextually reconstructed riddle is appended to the input. 
For a fair comparison, the number of shots refers to the total number of examples included in the prompt, regardless of whether they are new riddles retrieved from the dataset or contextually reconstructed examples.
Thus, for example RISCORE with 4 shots has the same 2 examples used in the 2-shot \textbf{FS}  technique augmented with the 2 contextual reconstructed samples of them; ultimately, RISCORE keeps only half of the examples drawn from the original dataset, while the remainder consists of reconstructed examples. Therefore, the 2, 4, and 8-shot RISCORE settings correspond to 1, 2, and 4 exemplars drawn from the dataset respectively, with the rest of the demonstrations being their context reconstructions. 

We primarily utilize Llama3-8B and Llama3-70B to generate both Question-Answer pairs and distractors. The Question-Answer pairs are produced in both \textbf{ZS} and \textbf{FS} settings. More details regarding these techniques can be found in App. \ref{sec:prompts}.



Lastly, in our experiments, we differentiate between RISCORE\textsubscript{m}, which utilizes manually created reconstructions (when available, as in existing BrainTeaset context reconstructions), and RISCORE, which employs our fully automated method for generating reconstructions.
The prompt structure for the system-user interaction in this method is the same as in the \textbf{FS} method outlined in App. \ref{sec:fewshot}. The sole distinction is in the examples provided.

\begin{table*}[t!]  
    \centering
    \begin{minipage}{0.45\textwidth}

    \centering
    \small
    \begin{tabular}{P{1.4cm}|c|P{0.5cm}P{0.5cm}P{0.5cm}P{0.5cm}P{0.5cm}}
    \midrule
  \textbf{Method}  & \textbf{N.} & 
    \rotatebox{90}{\textbf{Llama3-70B}} & \rotatebox{90}{\textbf{Mistral-8x7B}} & \rotatebox{90}{\textbf{Llama3-8B}} & \rotatebox{90}{\textbf{Mistral-7B}} & \rotatebox{90}{\textbf{Qwen2-7B}}\\
    \midrule

    CoT ZS & 0 & 0.725 & 0.550 & 0.633 & 0.450 & 0.458 \\

    \midrule
        \multicolumn{7}{c}{\textbf{Randomly Selected Shots}} \\
    \midrule

     & 2 & 0.758 & 0.617 & 0.633 & 0.475 & 0.608 \\
    CoT FS & 4 & 0.683 & 0.583 & 0.608 & 0.508 & 0.650 \\
     & 8 & 0.708 & 0.642 & 0.658 & 0.508 & 0.667 \\

     \cmidrule(lr){1-7}

     & 2 & 0.775 & 0.617 & 0.633 & 0.517 & 0.642 \\
    FS Rand & 4 & 0.808 & 0.683 & 0.642 & 0.483 & 0.608 \\
     & 8 & 0.775 & 0.617 & 0.675 & 0.483 & 0.642 \\
    
     \cmidrule(lr){1-7}

     & 2 & 0.783 & 0.625 & 0.667 & 0.458 & 0.608 \\
    RISCORE\textsubscript{m} & 4 & 0.758 & 0.617 & 0.675 & 0.517 & 0.625 \\
     & 8 & 0.800 & 0.650 & 0.667 & 0.400 & 0.592 \\

    \midrule
        \multicolumn{7}{c}{\textbf{Semantically Similar Shots}} \\
    \midrule
     & 2 & \underline{0.825} & \underline{0.692} & 0.700 & 0.517 & 0.600 \\
    FS Sim & 4 & 0.792 & 0.683 & 0.717 & 0.458 & 0.633 \\
     & 8 & 0.783 & 0.667 & \underline{0.767} & \underline{0.533} & \underline{0.650} \\
    
     \cmidrule(lr){1-7}

     & 2 & 0.783 & 0.675 & \textbf{0.767} & 0.483 & 0.667 \\
    RISCORE\textsubscript{m} & 4 & \textbf{0.833} & 0.708 & 0.742 & \textbf{0.567} & 0.642 \\
     & 8 & 0.808 & \textbf{0.708} & 0.758 & 0.550 & \textbf{0.667} \\
    
    \bottomrule

    \end{tabular}

    \caption{Model performance for \textit{BrainTeaser} using baselines and RISCORE\textsubscript{m} prompting. The best \textbf{FS} results are \underline{underlined}, while best overall results per model are highlighted in \textbf{bold}. More detailed results can be found in Tables \ref{tab:sentence_results1} and \ref{tab:sentence_results2}.}
    \label{tab:brainteaser_main}

    \end{minipage}
    \hfill
    \begin{minipage}{0.45\textwidth}

    \centering
    \small
    \begin{tabular}{P{1.4cm}|c|P{0.5cm}P{0.5cm}P{0.5cm}P{0.5cm}P{0.5cm}}
    \midrule
  \textbf{\rotatebox{0}{Method}} &  \textbf{N.} & 
    \rotatebox{90}{\textbf{Llama3-70B}} & \rotatebox{90}{\textbf{Mistral-8x7B}} & \rotatebox{90}{\textbf{Llama3-8B}} & \rotatebox{90}{\textbf{Mistral-7B}} & \rotatebox{90}{\textbf{Qwen2-7B}}\\
    \midrule
        \multicolumn{7}{c}{Llama3-70B ZS  for QA \& Llama3-8B distractors} \\
    \midrule

     & 2 & 
     0.792 & 0.667 & 0.625 & 0.492 & \textbf{\underline{0.625}}\\

      RISCORE & 4 & 
     \textbf{\underline{0.792}} & 0.642 & 0.675 & \textbf{\underline{0.467}} & 0.625\\

       & 8 & 
     \textbf{\underline{0.808}} & \textbf{\underline{0.683}} & 0.700 & 0.475 &  0.642 \\
     \midrule

      \multicolumn{7}{c}{Llama3-70B FS for QA \& Llama3-8B distractors} \\
    \midrule 
    
      & 2 & 
     0.750 & 0.675 & 0.683 & 0.475 & \textbf{\underline{0.625}}\\

      RISCORE & 4 & 
     \textbf{\underline{0.792}} & 0.650 & 0.658 & \textbf{\underline{0.558}} & \textbf{\underline{0.658}}\\

       & 8 & 
     \textbf{\underline{0.808}} & \textbf{\underline{0.675}} & 0.742 & 0.517 &  \textbf{\underline{0.658}} \\
     \midrule
    \multicolumn{7}{c}{Llama3-70B FS for QA \& Llama3-70B distractors} \\
    \midrule 
   
     & 2 & 
     0.783 & 0.667 & 0.683 & 0.500 & \textbf{\underline{0.617}}\\

      RISCORE & 4 & 
     \textbf{\underline{0.792}} & 0.642 & 0.667 & \textbf{\underline{0.508}} & 0.617\\

       & 8 & 
     0.767 & \textbf{\underline{0.683}} & 0.700 & 0.500 & 0.617 \\

    \bottomrule

    \end{tabular}

    \caption{Model performance for \textit{BrainTeaser} using RISCORE prompting. Similarity-based selection was employed for choosing all the exemplars. Results that surpass the \textbf{FS} method with semantically similar examples (\textbf{FS Sim}, check Table \ref{tab:brainteaser_main}), using the same number of shots, are \textbf{\underline{underlined}}. More detailed results can be found in Tables \ref{tab:sentence_results2} and \ref{tab:sentence_results3}.}
    \label{tab:brainteaser_ours}
    \end{minipage}
\end{table*}

\subsection{Models}
To this end, we test various models of different scales on their reasoning abilities. Specifically, we examine Llama3 with 8 and 70 billion parameters respectively \cite{llama3modelcard}, Mistral-7B and Mistral-8x7B \cite{jiang2023mistral7b}, as well as Qwen2-7B \cite{yang2024qwen2technicalreport}. This diverse set of models allows us to investigate the impact of contextually reconstructed examples on models with varying sizes and parameterizations.  By treating each model as a black-box (we merely prompt them and gather their responses), RISCORE adapts seamlessly to both open-source and proprietary models, demonstrating its versatility across a wide range of LLMs.
Further details regarding the model selection process and technical specifications can be found in App. \ref{sec:models}.

\section{Results}
\label{sec:results}
In this section, we present and analyze the results of our experiments. Additional results utilizing different metrics and hyperparameter settings are available in App. \ref{sec:full_results}.

\subsection{BrainTeaser results}
Table \ref{tab:brainteaser_main} presents the performance outcomes of various prompting techniques applied to BrainTeaser data. 
A key observation is the underperformance of the \textbf{CoT FS} method relative to \textbf{FS} techniques, even when examples are supplemented with manually crafted explanations. This underperformance persists regardless of the model size. As expected \cite{dong2024surveyincontextlearning}, exemplar selection based on semantic similarity (\textbf{FS Sim}) consistently improves results when chosen over random selection (\textbf{FS Rand}), highlighting the value of semantic relevance to shot selection. This is a significant reference point for RISCORE, since a suitable initial exemplar selection is critical in properly guiding model reasoning when enhanced with contextually reconstructed examples.

Importantly, RISCORE\textsubscript{m} consistently outperforms \textbf{FS Sim} across all 4-shot and 8-shot configurations, as shown in Table \ref{tab:brainteaser_main}. For instance, when comparing models with the same number of examples—such as 4-shot \textbf{FS Sim} versus 2+2-shot RISCORE\textsubscript{m}, or 8-shot \textbf{FS Sim} versus 4+4-shot RISCORE\textsubscript{m}—RISCORE\textsubscript{m} demonstrates superior performance. The advantages of RISCORE\textsubscript{m} are particularly evident in its ability to maintain robust reasoning.

To better understand this advantage, we can examine the ability of RISCORE\textsubscript{m} to mitigate noise from suboptimal shot selections, a notable issue in the \textbf{FS Sim} results. For example, with the Llama3-70B model, the 2-shot \textbf{FS Sim} achieves a performance score of 0.825. However, adding two more semantically similar examples reduces the score to 0.792. In contrast, the 2+2-shot RISCORE\textsubscript{m} achieves a score of 0.833—a 4\% improvement over \textbf{FS Sim} with the same number of examples (vs 4-shot). This trend is consistent across other models. Similarly, with the smaller Mistral-7B model, the 2-shot \textbf{FS Sim} score is 0.517, but adding two more examples reduces it to 0.458. In contrast, the 2+2-shot RISCORE\textsubscript{m}, using contextually reconstructed examples, achieves a score of 0.567, outperforming \textbf{FS Sim} under identical shot conditions (vs 4-shot). These results indicate that RISCORE\textsubscript{m} effectively mitigates the noise introduced by suboptimal shot selections.

\paragraph{Using the automated method}

The manually curated RISCORE\textsubscript{m} results serve as a benchmark, showcasing the potential upper bound of performance gains achievable with high-quality, handpicked riddles and distractors. Despite this, the automated RISCORE method consistently enhances performance by effectively leveraging reconstructed contexts, as shown in  Table \ref{tab:brainteaser_ours}. While the improvements using the automated RISCORE are understandably smaller than those achieved with manually curated examples (check RISCORE\textsubscript{m} results of Table \ref{tab:brainteaser_main}), they are nonetheless consistent and significant across all models tested. For example, Llama3-70B demonstrates significant improvement when transitioning from \textbf{FS Sim} with 8 semantically similar shots (0.783) to the RISCORE-augmented setting with 8 shots (0.808), which incorporates the original examples and their automated reconstructions. This trend is evident across all relatively smaller models, with noticeable performance gains. For Qwen2-7B, a 2-shot and 4-shot RISCORE setup results in a 2.5\% improvement compared to the same-shot \textbf{FS Sim} setting, while for Mistral-7B, the improvement is even more pronounced—rising by up to 10\% from a baseline of 0.458 under certain conditions in the 4-shot setting.


These results underscore the value of context reconstruction, particularly when the number of shots is limited, demonstrating that automated reconstruction can effectively complement semantically chosen examples.
Interestingly, the results reveal that in most cases, RISCORE's approach of combining N/2 semantically selected examples with their automated reconstructions outperforms the mere placement of the same number of examples retrieved from the dataset (indicated by the underlined results). 
However, there are instances where \textbf{FS Sim} outperforms RISCORE, primarily due to the selection constraints of RISCORE. In cases where the semantic similarity algorithm identifies an optimal example ranked lower than N/2 in a full \textbf{FS} setup, RISCORE may miss these shots due to its focus on N/2 examples plus reconstructions. This limitation indicates that RISCORE’s performance can be constrained by the quality of initial semantic similarity rankings and the resulting shot selection.

\begin{table*}[!htbp]  
    \centering
    \begin{minipage}{0.45\textwidth}
       
    \centering
    \small
    \begin{tabular}{c|c|P{0.65cm}P{0.65cm}P{0.65cm}P{0.65cm}P{0.65cm}}
    \midrule
  \textbf{Method} &  \textbf{N.} & 
    \rotatebox{90}{\textbf{Llama3-70B}} & \rotatebox{90}{\textbf{Mistral-8x7B}} & \rotatebox{90}{\textbf{Llama3-8B}} & \rotatebox{90}{\textbf{Mistral-7B}} & \rotatebox{90}{\textbf{Qwen2-7B}}\\
    \midrule

    CoT ZS & 0 & 0.775 & 0.675 & 0.619 & 0.589 & 0.608 \\

    \midrule
        \multicolumn{7}{c}{\textbf{Randomly Selected Shots}} \\
    \midrule
    
       & 2 & 0.789 & 0.692 & 0.625 & 0.594 & 0.667 \\
      CoT FS & 4 & 0.783 & 0.686 & 0.672 & 0.603 & 0.656 \\
       & 8 & 0.783 & 0.697 & 0.658 & 0.597 & 0.625 \\

     \cmidrule(lr){1-7}

       & 2 & 0.769 & 0.706 & 0.672 & 0.586 & 0.689 \\
      FS Rand & 4 & 0.772 & 0.719 & 0.639 & 0.586 & 0.683 \\
       & 8 & 0.800 & 0.711 & 0.672 & 0.586 & 0.700 \\

    \midrule
        \multicolumn{7}{c}{\textbf{Semantically Similar Shots}} \\
    \midrule
       & 2 & 0.792 & \textbf{0.714} & 0.706 & 0.608 & 0.722 \\
      FS Sim & 4 & \textbf{0.817} & 0.692 & \textbf{0.711} & \textbf{0.633} & 0.714 \\
       & 8 & 0.800 & 0.675 & 0.681 & 0.611 & \textbf{0.731} \\
    
    \bottomrule

    \end{tabular}

    \caption{Model Performance for \textit{RiddleSense} using baseline techniques. The best
results overall are in \textbf{bold}. More detailed results can be found in Table \ref{tab:rs_results}. Note that no RISCORE\textsubscript{m} numbers are reported, since \textit{RiddleSense} does not contain any ground truth reconstructed riddles.}
    \label{tab:rs_main}

    \end{minipage}
    \hfill
    \begin{minipage}{0.45\textwidth}
       
    \centering
    \small
    \begin{tabular}{c|c|P{0.55cm}P{0.55cm}P{0.55cm}P{0.55cm}P{0.55cm}}
    \midrule
  \textbf{\rotatebox{0}{Method}} &  \textbf{N.} & 
    \rotatebox{90}{\textbf{Llama3-70B}} & \rotatebox{90}{\textbf{Mistral-8x7B}} & \rotatebox{90}{\textbf{Llama3-8B}} & \rotatebox{90}{\textbf{Mistral-7B}} & \rotatebox{90}{\textbf{Qwen2-7B}}\\
    \midrule
    \multicolumn{7}{c}{Llama3-70B fewshot for QA \& Llama3-70B distractors} \\
    \midrule

     & 2 & \textbf{\underline{0.792}} & 0.672 & 0.692 & 0.600 & 0.697 \\
    RISCORE & 4 & 0.783 & 0.689 & \textbf{\underline{0.722}} & 0.600 & \textbf{\underline{0.717}} \\
     & 8 & 0.789 & \textbf{\underline{0.700}} & \textbf{\underline{0.708}} & 0.597 & \textbf{\underline{0.731}} \\

     \midrule

      \multicolumn{7}{c}{Llama3-70B fewshot for QA \& Llama3-8B distractors} \\
    \midrule 
    
     & 2 & 0.786 & \textbf{\underline{0.719}} & 0.681 & 0.603 & 0.681 \\
    RISCORE & 4 & 0.789 & 0.686 & 0.686 & 0.606 & 0.697 \\
     & 8 & 0.775 & \textbf{\underline{0.689}} & \textbf{\underline{0.706}} & \textbf{\underline{0.617}} & 0.719 \\
    
     \midrule
        \multicolumn{7}{c}{Llama3-8B zeroshot  for QA \& Llama3-8B distractors} \\
    \midrule 
   
     & 2 & \textbf{\underline{0.792}} & 0.681 & 0.689 & 0.589 & 0.694 \\
    RISCORE & 4 & 0.778 & \textbf{\underline{0.714}} & 0.700 & 0.600 & 0.683 \\
     & 8 & \textbf{\underline{0.806}} & \textbf{\underline{0.689}} & \textbf{\underline{0.686}} & \textbf{\underline{0.614}} & 0.689 \\
     
    \bottomrule

    \end{tabular}

    \caption{Model performance for \textit{RiddleSense} using RISCORE prompting. Similarity-based selection was employed for choosing all the exemplars. Results that surpass the FS method with semantically similar examples, using the same number of shots, are \textbf{\underline{underlined}}. More detailed results can be found in Table \ref{tab:rs_results1}.}
    \label{tab:rs_ours}
    \end{minipage}
\end{table*}

\subsection{RiddleSense}
In this dataset, RISCORE can only be applied to automatically generated examples because the dataset's format lacks context reconstructions for its questions.

Table \ref{tab:rs_main} presents the results of the baseline techniques using various models in the RiddleSense dataset. Once again, the results confirm that the few-shot technique, utilizing semantically similar exemplars for in-context learning, consistently delivers the best performance across all tested models.

In Table \ref{tab:rs_ours}, we present the RiddleSense results using the proposed method for context reconstruction of each input. A clear trend emerges when comparing a simple 8-shot exemplar selection based on semantic similarity with the 8-shot setting of our method, where the top 4 semantically similar examples are augmented with our generated contextual pairs. Notably, the results show that our method consistently outperforms the standard 8-shot exemplar approach, demonstrating a significant improvement in model performance across various instances. An example of this trend is observed with the Llama3-8B model, where our method scores 0.708, approximately 2\% higher than the few-shot setting based on semantic similarity, which achieves a score of 0.681. In the case of Mistral-8x7B, RISCORE avhieves a score of 0.700, a 2.5\% gain over \textbf{FS Sim}’s 0.675, further demonstrating its consistent advantage. 
The same pattern is evident when comparing the two 4-shot settings. We achieve similar or marginally better accuracy using a total of just 4 examples—two original and two generated contextual reconstructions. 
This underlines the effectiveness of integrating contextually reconstructed pairs in enhancing model accuracy. 
We do not achieve a significant performance boost; however, we attain similar or marginally better results while relying on less grounded knowledge. This demonstrates the efficiency of our method, as it maintains comparable performance with fewer, yet strategically selected, exemplars, shifting the focus towards reasoning relevance of exemplars rather than quantity.




\subsection{Quality of context reconstructed riddles}

To generate contextually reconstructed riddles, we use Llama3 models with 8 billion and 70 billion parameters in both \textbf{FS} and \textbf{ZS} settings. We find that the Llama3-8B model struggles to produce high-quality Question-Answer pairs for the BrainTeaser dataset and is therefore not suitable for RISCORE. This difficulty likely arises from the BrainTeaser dataset's demand for lateral thinking, which is particularly challenging for smaller reasoners.
The Question-Answer pairs are essential, and if their quality is insufficient—as observed when leveraging Llama3-8B on BrainTeaser—high-quality distractors cannot compensate for this deficiency on their own. To address this issue, our preprocessing and filtering process ensures that only high-quality contextual examples are retained, maintaining the effectiveness of the method without being compromised by low-quality generations.
However, for vertical thinking riddles, the smaller model effectively generates riddles even in the \textbf{ZS} setting, which, when used in the \textbf{FS} setting, can lead to increased performance compared to real examples drawn from the dataset.

\section{Conclusion}

We explore the riddle-solving capabilities of LLMs through multiple-choice formats, investigating how different prompts affect riddle performance requiring diverse reasoning. Our findings suggest that in few-shot settings, augmenting exemplars with contextually reconstructed examples improves performance. We introduce a novel prompting method, RISCORE, validated using the BrainTeaser dataset featuring manually crafted contextual riddles. Inspired by these results, we provide a method for automatically generating contextual reconstructions for multiple-choice riddles, demonstrating that RISCORE enhances LLMs' abilities in lateral and vertical thinking tasks.


\section*{Limitations} This method has some limitations. For instance, the placement of exemplars is based on semantic similarity, although cues unrelated semantically might be more crucial for reasoning, such as reasoning path similarity or diversity \cite{zhang2022automaticchainthoughtprompting}. To address this, future work will employ more advanced methods for selecting exemplars for few-shot settings, like those suggested \cite{sultan-shahaf-2022-life}. Additionally, the current study, having tested only two datasets, limits the broader impact of the conclusions since outcomes might vary with other datasets. Future steps will include testing the RISCORE prompting method with additional datasets across various tasks \cite{zhang2022birdqabilingualdatasetquestion}.
Finally, our experimentation is exclusively focused on the English language, therefore we cannot conclude whether our findings are applicable on other languages.

\section*{Ethical Considerations} There are no considerable risks associated with our work.

\section*{Acknowledgments} The research work was supported by the Hellenic Foundation for Research and Innovation (HFRI) under the 3rd Call for HFRI PhD Fellowships (Fellowship Number 5537).

\bibliography{custom}

\appendix

\section{Question-Answer Pairs Generation}
\subsection{Process details}
\label{sec:qa_construction_det}
This section of our pipeline involves several steps, and subsequent steps after the initial one vary based on the dataset format.

Initially, we need to create contextual reconstructed Question-Answer pairs, temporarily disregarding distractors for both datasets. To generate these Question-Answer pairs, we can leverage the capabilities of large language models (LLMs) in two ways.
First, we use a system-user prompt template to provide specific instructions to the model, guiding it through the process of producing a new Question-Answer pair. This approach maintains the underlying commonsense premise of the original question while altering both the question and the answer to fit a new situational context. This method operates in a zero-shot setting, meaning the model generates the new pairs without prior examples or training specific to the task.
The second approach is quite similar but leverages the pre-existing pairs of original and contextually reconstructed questions from BrainTeaser. In this few-shot setting, we use the same instructions as in the zero-shot approach, but we enhance the model's understanding by providing in-context examples. Specifically, we include a pair consisting of an original Question-Answer and its corresponding contextually reconstructed version. This helps the model better grasp the task at hand. It is important to note that these few-shot examples are of high quality, given their manual construction and careful curation.
In the few-shot approach, the provided examples are static. It is important to highlight that, in BrainTeaser, when selecting examples for few-shot exemplars from the training dataset, we ensure that the examples chosen are not selected for reconstruction. This avoids any overlap between the few-shot exemplars and the examples for which we are requesting a context change, ensuring that the given pair examples are distinct from the requested ones.
The corresponding system and user prompts regarding the above tasks can be found in App. \ref{sec:qa_construction}.
\paragraph{Models utilized}
In this approach, we experiment with two models to assess the impact of model size on the quality of contextual reconstructions and the overall performance across both datasets. We use a relatively small model, Llama3-8B \cite{llama3modelcard}, and a larger model from the same family, Llama3-70B \cite{llama3modelcard}. These models were selected due to their strong performance across a diverse range of tasks, outperforming other models of comparable size. By incorporating both a smaller model, Llama3-8B, and a larger model, Llama3-70B, we aim to systematically investigate the effect of model size on the quality of contextual reconstructions and its overall impact on performance. This exploration allows us to gain deeper insights into how model capacity influences the effectiveness of the generated examples across different datasets.

\paragraph{Quality Assurance for Generated Text}
After generating the contextual reconstructed Question-Answer pairs for all initially selected examples from the training set, we perform a small but necessary quality control to ensure that the generated pairs adhere to several high-level criteria.
For the BrainTeaser dataset, the filtering process is more flexible, as the questions and answers can vary in length. The only constraint is that the questions must have a minimum length of 7 words, while the answers can be as long as necessary. As a result, the number of generated pairs filtered out is relatively low, around 2\%, ensuring that the majority of reconstructions meet the quality criteria.
In contrast, the RiddleSense dataset requires stricter limitations due to its typically shorter answers. For this dataset, questions must be at least 6 words in length, while answers must not exceed 7 words, reflecting the dataset's concise nature. As a result of these stricter rules, approximately 10\% of the generated reconstructions for the RiddleSense dataset are filtered out for not meeting the required standards.
It is important to note that the minimum question length requirement is implemented to ensure that the generated questions maintain a sufficient level of complexity and depth. Questions with fewer than 6 or 7 words are unlikely to provide the necessary detail to establish a well-structured and coherent riddle. This limitation helps preserve the quality and integrity of the riddles by ensuring that they are adequately framed to challenge and engage the model effectively.

\subsection{Prompts}
\label{sec:qa_construction}
First, we will present the \textbf{ZS} prompt configuration used for generating the contextually reconstructed Question-Answer pairs.
\par\noindent\rule{0.5\textwidth}{0.4pt}
\small \textbf{System Prompt:}\\
You are an expert in context reconstruction. Your task is to receive a question along with its correct answer and adapt them to a new scenario while maintaining the misleading commonsense premise.\\
\\
Please follow these steps:\\
\\
- First, you will receive an unsolved riddle along with five answer options. Analyze the given setting and identify the connection between the question and its correct answer.\\
\\
- Modify the original question and correct answer to fit a different situational context, ensuring that the underlying logic and relationship between them are preserved.\\
\\
- Ensure that both the new question and the new correct answer are distinct from the originals.\\
\\
\textbf{User Prompt:}\\
\begin{verbatim}
Question: ```{QUESTION}```

Correct answer: ```{ANSWER}```\end{verbatim}
\par\noindent\rule{0.5\textwidth}{0.4pt}
\normalsize 

Next, with minor modifications, we introduce the \textbf{FS} prompt configuration that follows the same logic but incorporates in-context examples to guide the model's responses.
\par\noindent\rule{0.5\textwidth}{0.4pt}
\small \textbf{System Prompt:}\\
You are an expert in context reconstruction. Your task is to receive a question along with its correct answer and adapt them to a new scenario while maintaining the misleading commonsense premise.\\
\\
Please follow these steps:\\
\\
First, review an example provided with its context reconstruction, which illustrates the type of transformation you will need to perform.\\
\\
Next, you will receive an unsolved riddle along with five answer options. Analyze the given setting and identify the connection between the question and its correct answer.\\
\\
Modify the original question and correct answer to fit a different situational context, ensuring that the underlying logic and relationship between them are preserved.\\
\\
Ensure that both the new question and the new correct answer are distinct from the originals.\\
\\
\textbf{User Prompt:}
\begin{verbatim}
{EXAMPLES}
\end{verbatim}
Adapt the following riddle - answer pair while taking into consideration the examples above regarding context reconstruction:
\begin{verbatim}
Question: ```{QUESTION}```

Correct answer: ```{ANSWER}```\end{verbatim}
\par\noindent\rule{0.5\textwidth}{0.4pt}
\normalsize 

In the above setting, it is understood that the values of the \textbf{EXAMPLES} represent pairs of answered Question-Answer examples: the original and its contextually reconstructed counterpart, both sourced from the BrainTeaser dataset.

\section{Distractor Creation}
\label{sec:ditractors}

\subsection{BrainTeaser}
\label{sec:dist_bt}
In this process, we leverage the capabilities of large language models (LLMs) to understand and rephrase context in order to generate distractors. We use two different pipelines to produce at least three distractors, ensuring coverage of various potential situations.
\circled{1} The first approach involves prompting the model with a system-user prompt that instructs it to analyze the given Question-Answer pair. The model is tasked with understanding the riddle, identifying the reasoning process that links the question to the answer, and then suggesting a distractor based on the more challenging or deceptive aspects of the concept. This approach yields one of the three required distractors.
\circled{2} In the second approach, the process is more intricate and can result in distractors of questionable quality. We prompt the model with a system-user prompt, providing it with the reconstructed question without its answer, along with the original question's incorrect distractors, after removing the option ``None of the above''.
The model is then tasked with modifying the concept of the given distractors by incorporating elements from the setting described in the question. Importantly, the correct answer is not provided to ensure that the generated distractors are distinctly different from the correct answer and effectively capture a varied interpretation of the question's context. Certainly, there are instances where the model generates sub-optimal contexts for the distractors provided. Despite these cases, the distractors remain incorrect and serve their purpose, though they may not always be sufficiently challenging. Our experiments indicate that when the model is given both the original and the contextual reconstructed examples, the minor issue of lower-quality distractors does not significantly impact overall performance. Now, we have also created two new distractors that are somewhat contextually relevant to the setting, adding an additional layer of coherence to the generated options. This ensures that the distractors are not only incorrect but also related to the underlying premise of the riddle, enhancing the overall quality of the multiple-choice options.
To create the final dataset, we randomly select two of the three generated distractors and shuffle them with the correct answer in random order. Finally, we append the option ``None of the above'' as the last choice. In cases where ``None of the above'' is the correct answer, we use all three generated distractors. With this approach, our dataset is prepared and ready for use.

\paragraph{Prompts used}
First, we will provide the system-user prompt used to task the model with understanding the riddle, identifying the reasoning process linking the question to the answer, and then suggesting a distractor based on the more challenging or deceptive aspects of the concept.
\par\noindent\rule{0.5\textwidth}{0.4pt}
\small \textbf{System Prompt:}\\
Your task is to act as a concept grasper. You will be given a riddle and its correct answer.\\
\\
Your goal is to understand the connection between the riddle and the correct answer, focusing on the tricky parts. Based on these tricky aspects, propose a plausible wrong answer that someone might give.\\ 
\\
The wrong answer should be short, concise, and limited to one sentence.\\
\\
- Riddle:\\
\\
- Correct Answer:\\
\\
Response format:\\
\\
- Wrong Answer:\\
\\
\textbf{User Prompt:}
\begin{verbatim}
- Riddle: {QUESTION}

- Correct Answer: {ANSWER}\end{verbatim}

\par\noindent\rule{0.5\textwidth}{0.4pt}
\normalsize 

Now we will provide the system-user prompt used for the second method.
\par\noindent\rule{0.5\textwidth}{0.4pt}
\small \textbf{System Prompt:}\\
You will be given a sentence without context and then provided with a specific context.\\
\\
Your task is to rewrite the sentence so that it aligns with the given context, while keeping it as close as possible to the original meaning.\\
\\
The purpose is to adapt the sentence to the context, not to answer any questions related to the context.\\
\\
- Sentence (out of context):\\
\\
- Context:\\
\\
Response format:\\
\\
- Sentence:\\
\\
\textbf{User Prompt:}
\begin{verbatim}
- Sentence (out of context): {ORI_CHOICE}

- Context:{QUESTION}\end{verbatim}

\par\noindent\rule{0.5\textwidth}{0.4pt}
\normalsize 

Here, the value of \textbf{ORI\_CHOICE} refers to the distractors from the original instance, excluding the "None of the above" option. We create prompts for the model with each of these distractors individually.

\subsection{RiddleSense}
\label{sec:dist_rs}
In this setting, the approach is fundamentally different. The answers and distractors in the original dataset are primarily one-word responses, while the questions feature detailed settings with punctuation, conjunctions, and more complex structures. To handle this, we also generate distractors using two distinct pipelines tailored to this format.

\paragraph{First pipeline} The first pipeline involves a more granular approach by splitting the contextual reconstructed question into subphrases based on punctuation or conjunctions. If this yields fewer than three distinct subphrases, we further split the sentence at the position of the word ``and'' to create additional segments. We also detect the presence of interrogative words, and if the question is purely descriptive without any direct question, we append the phrase ``What am I?'' at the end. This is not chosen arbitrarily but follows the standard structure of many riddles in this dataset, where "What am I?" is the common question leading to single-word answers.
Now that we have broken the riddle into sub-phrases and appended the appropriate question, we can prompt the model to generate a wrong answer for each sub-phrase concatenated with the question. This method ensures that the distractors align with different parts of the riddle's setting. However, since some sub-phrases may contain key ideas central to the riddle, the model may still generate answers too similar to the correct one. To address this issue, we incorporate an additional intermediate step utilizing zero-shot classification. Specifically, we use the \textit{facebook/bart-large-mnli} model \footnote{\href{https://huggingface.co/facebook/bart-large-mnli}{facebook/bart-large-mnli}} \cite{lewis2019bartdenoisingsequencetosequencepretraining} provided by Hugging Face. This model classifies the correct answer into one of eight general categories: \textit{'food', 'person', 'object', 'animal', 'nature', 'time', 'place', 'concept'}. These categories are chosen to be mutually exclusive to avoid overlap.
In our approach, we first predict the category of the correct answer using this model. We then leverage this classification to guide the generation of distractors. For each sub-phrase concatenated with the question, we provide the LLM with the two most similar categories (excluding the correct answer's category). The model is prompted with a system-user instruction to produce a correct answer of the given setting in the given category. This method ensures that the generated distractors are not only incorrect but also contextually aligned with the riddle, thus maintaining the challenge for the model. After applying filtering to ensure the distractors are distinct and relevant, we have produced several distractors. The LLMs utilized in this pipeline include the previously mentioned Llama3-8B \cite{llama3modelcard} and its larger counterpart, Llama3-70B \cite{llama3modelcard}.
\paragraph{Second pipeline} In cases where the distractors produced by the above pipeline are insufficient, we use WordNet \cite{wordnet2006} to augment our distractor set. For each generated distractor, or if necessary, for the distractors from the original question, we retrieve synonyms and hyponyms from WordNet. These additional terms are then included as potential distractors. After compiling these distractors, we randomly select four of them and add the correct answer in a random order to complete the set. This approach ensures that we have a diverse and comprehensive set of distractors for each question. To ensure the quality of our distractors, we impose a limitation that at least two of the required four distractors must be generated using the first approach. 
This is due to the fact that the distractors generated through WordNet augmentation tend to be of inferior quality compared to those produced directly by the model.
If this requirement is not met, we skip the particular train set instance for producing a contextual reconstruction. This approach helps maintain the overall quality and relevance of the distractors in our dataset.

\paragraph{Prompts used}
We will present the system-user prompt configuration used in the first pipeline. In this approach, for each sub-phrase concatenated with the question, we provide the model with two categories, excluding the correct answer's category. For each category, a separate system-user prompt is issued to instruct the model to generate a correct answer that fits the given setting within the specified category.
\par\noindent\rule{0.5\textwidth}{0.4pt}
\small \textbf{System Prompt:}\\
Task: Provide a concise, relevant answer to the given question within the specified category.\\
\\
Constraints:\\
- The answer should not exceed three words.\\
- Follow the exact format provided below.\\
\\
Response Format:\\
\\
Answer: ...\\
\\
\textbf{User Prompt:}
\begin{verbatim}
Question:  ```{QUESTION}``` 

Category: {CATEGORY}\end{verbatim}

\par\noindent\rule{0.5\textwidth}{0.4pt}
\normalsize 

\section{Pairs Prompting}
\label{sec:pairs_prompting}
We have successfully created the desired contextual reconstructions for both datasets. However, an issue remains: some of the originally selected examples do not have corresponding reconstructed examples due to the quality control filtering process.

To address this issue, we first use the original most semantically similar examples as in-context learning exemplars, appending their corresponding automatically generated contextual pairs. If these exemplars are not sufficient for the required settings of two, four, or eight (i.e., \textbf{RISCORE}), we employ a more structured approach rather than adding examples randomly. \circled{1} We begin by generating embeddings for the set of original examples and their contextual reconstructions that have not already been included in the current exemplars. Using cosine similarity, we then identify the most similar examples from this set. Importantly, these similar examples might not be part of the original training set but could be among the reconstructed examples.
\circled{2} We select the most similar examples and pair each with its corresponding pair, ensuring that we add either the original or the reconstructed example as needed. This process is repeated until the number of exemplars meets the required quantity.

\section{Dataset Description}
\label{sec:dataset}

In this work we are using two different datasets of multiple QA format. \circled{1} The first one is the BrainTeaser dataset, which consists of two sub-tasks regarding both lateral thinking challenges.
In addition to the original puzzles, the dataset includes adversarial subsets created by manually modifying the original BrainTeaser while preserving their reasoning paths. The original data were perturbed in two ways: First, there is a \textit{semantic reconstruction} of each original question, ensuring that the answers and distractors remain unchanged. Second, the original data undergoes \textit{context reconstruction}, where the reasoning path is preserved, but the brain teaser is rephrased to describe a new situational context. Each question offers four possible choices, with the last option always being ``None of the above''. It is important to note that our experiments are focused exclusively on the sentence puzzle (SP) setting.
\circled{2} The second dataset used in our experiments is RiddleSense. This dataset features riddle-style questions that require sophisticated commonsense reasoning and a strong grasp of figurative language to answer accurately. It is structured as a multiple-choice question answering task, with riddles that test the model’s capacity to navigate and interpret nuanced commonsense scenarios. The reasoning challenges posed by RiddleSense are closely aligned with those found in BrainTeaser.
RiddleSense offers one correct answer and four distractors for each riddle. This contrasts with BrainTeaser, which includes a ``None of the above'' option as the final choice, thereby introducing a more challenging setting.


\subsection{Data statistics}
\paragraph{Format of Datasets}
The BrainTeaser dataset is divided into three splits: train, development, and a hidden test set used for evaluation. Although RiddleSense follows a similar format, we did not have direct access to its hidden test set. As a result, we utilized the development data split of RiddleSense for our experiments.

\paragraph{Prepossessing and filtering}
Due to data leakage across the two datasets, we were required to exclude semantically similar examples from the test split utilized. This leakage is understandable, as the sources for English riddles are limited, and the number of unique riddles is finite. After identifying semantically similar riddles, we retained only one instance within the datasets, prioritizing BrainTeaser due to the higher quality of its questions and distractors, which were manually crafted.
Additionally, hardware limitations made processing the full RiddleSense  unfeasible. To manage this, we split its test set into two halves. To ensure the fairness and accuracy of this division, we evaluated various techniques on both halves. Following the completion of these procedures, we present the dataset statistics in the table below. Statistics regarding the datasets are provided in Table ~\ref{tab:data_stats}. More information regarding this process can be found in \ref{subsec:data_prepr}.

    

    

\begin{table}[ht!]
    \centering
    \small
    \resizebox{\columnwidth}{!}{%
    \begin{tabular} {l|l|l|l}
    \midrule
    \textbf{Dataset} & \textbf{Train} & \textbf{Dev} & \textbf{Test} \\
    \midrule
    BrainTeaser - SP & 507$_{(169 x 3)}$ & 120$_{(40 x 3)}$ & 120$_{(40 x 3)}$ \\
    \midrule
    
    RiddleSense (initial)&  & 1021 &  \\
    RiddleSense (filtered) & 3510 & 720 & --- \\
    RiddleSense (sampled 50\%) &  & 360 &  \\
    \midrule

    
    \end{tabular}%
    }
    \caption{Data statistics}
    \label{tab:data_stats}
\end{table}

\subsection{Evaluation metrics}
Given that our task is formatted as a multiple-choice question-answering problem, the evaluation metric employed will be accuracy. This metric will provide a measure of the proportion of correctly answered questions out of the total number of questions presented. Although accuracy is a straightforward metric, it is effective in providing useful insights and serves as a reliable measure for evaluation in this context.

\circled{1} For the BrainTeaser task, we not only track the overall accuracy but also monitor the accuracy for each specific type of instance, including original, semantic, and context reconstructions. This detailed tracking is feasible because the data in each set are balanced, with each original instance having corresponding semantic and context variations. This approach allows us to evaluate the model's proficiency in handling the same reasoning path under different conditions. The total accuracy is then calculated as the mean of these three individual metrics. 
We also track group-based accuracy, where a "group" refers to either two questions (original and semantically reconstructed) or three questions (original, semantically reconstructed and contextually reconstructed). This metric assesses the model's performance when all instances within a group are answered correctly. Group-based accuracy provides a broader perspective on the model's ability to handle these lateral thinking challenges across different types of question reconstructions. This format of evaluation metrics was both provided and requested by the creators of the dataset in an open Kaggle competition, and we are adopting it for our analysis. 

\circled{2} Unfortunately, the simple data row format in RiddleSense restricts our ability to track more detailed metrics comprehensively, limiting us to evaluating only the overall accuracy for this dataset.

\subsection{Dataset preprossessing}
\label{subsec:data_prepr}
Our goal in this process is to ensure that no question from the RiddleSense test set is present in the combined train, dev, and test datasets of BrainTeaser. To achieve this, we are using the Sentence Transformers library \footnote{\href{https://www.sbert.net/}{SentenceTransformers}} \cite{reimers-2019-sentence-bert} with the \textit{gte-large-en-v1.5} \footnote{\href{https://huggingface.co/Alibaba-NLP/gte-large-en-v1.5}{Alibaba-NLP/gte-large-en-v1.5}} \cite{zhang2024mgtegeneralizedlongcontexttext} model to generate text embeddings and perform semantic similarity comparisons. We convert each question from the entire BrainTeaser puzzle into embeddings that represent the semantic meaning of each question.
We then calculate the cosine similarity between these embeddings and those in the RiddleSense test set to identify duplicates or highly similar questions. A lower threshold is applied to define similarity, meaning that any question with a cosine similarity above this threshold is considered a near-duplicate and excluded from the test set.
After conducting several experiments, we determined that a cosine similarity threshold of 0.9 was optimal. This threshold is a delicate balance, as our questions are riddle-style and often rely on abstract or metaphorical language, making it challenging to clearly assess whether two questions are similar enough to warrant exclusion. Since riddles are not always straightforward in their wording, the threshold needed to be fine-tuned to avoid excluding questions that are conceptually distinct but might have overlapping language, ensuring a fair evaluation without overly aggressive filtering.
This means that any question from the RiddleSense test set that had a similarity score of 0.9 or higher to a question in the BrainTeaser datasets was removed to avoid redundancy. As previously mentioned, because we prioritize BrainTeaser questions due to their superior quality from manual annotation, we decided to remove any similar questions from the RiddleSense test set. This approach helps preserve the integrity of the evaluation process and ensures that the test set contains only unique and distinct questions, minimizing potential data leakage and improving the robustness of our results. Additionally, by removing these instances, we can perform more experiments on the remaining unique questions, optimizing our use of hardware resources and potentially yielding more comprehensive insights.

\section{Prompting details}
\label{sec:prompts}
\paragraph{Zero-shot Chain of Thought Prompting Technique}
This is the most straightforward prompting technique. In this approach, each instance from the test set of each dataset is presented to the model using a specific system-user prompt format. This technique operates in a zero-shot setting, meaning the model is not given any prior examples or training specific to the task at hand. Instead, it relies solely on the system prompt and user prompt to generate responses based on its pre-existing knowledge. The chaoin of thought character is a result of the usage of the phrase \textit{Let’s think step by step} in the end of each system prompt. The Chain of Thought (CoT) character is achieved by including the phrase \textit{“Let’s think step by step”} at the end of each system prompt. This approach aligns with the findings of \cite{kojima2023largelanguagemodelszeroshot}, which demonstrate that large language models can perform effectively as zero-shot reasoners by incorporating this specific phrase before presenting an unanswered test instance. The following are the system-user prompts that were used:
\par\noindent\rule{0.5\textwidth}{0.4pt}
\small \textbf{System Prompt:}\\
You will encounter a riddle that requires analytical thinking and reasoning to solve.\\
A riddle is a question or statement intentionally phrased so as to require ingenuity in ascertaining its answer or meaning, typically presented as a game.\\

Different ideas can be used in these riddles:\\
    1. Riddles often employ misdirection, leading you away from the actual solution.\\
    2. They include elements with double meanings, requiring a keen eye for words with dual interpretations.\\
    3. Metaphorical wordplay adds another layer, urging you to decipher figurative language.\\
    4. Look out for exaggeration, as riddles may present overly dramatic details to divert your attention.\\
    5. Common phrases and sayings may hide within the puzzle, demanding familiarity.\\
    6. Associations and irony play a crucial role, introducing unexpected connections.\\
    7. Numerical puzzles can also be part of the mystery, requiring you to decode their significance.\\
    8. Elemental imagery, drawn from nature, might hold key descriptors.\\
    9. Rhyming and sound clues can add a poetic dimension.
    10. Word Puzzles: Pay attention to anagrams, acrostics, and other wordplay elements.\\
    11. Also, it is important to note you should decode the upcoming riddle using everyday logic and creativity.\\
    \\
Approach the riddle with everyday logic and creativity, avoiding supernatural explanations.\\
You will be given an unsolved riddle and five options to choose the answer amongst them. Let's think step by step.\\
\\
\textbf{User Prompt:}
\begin{verbatim}
Riddle: ```
{RIDDLE}
```

Options:
[option 1]: ```{OPTION_1}```
[option 2]: ```{OPTION_2}```
[option 3]: ```{OPTION_3}```
[option 4]: ```{OPTION_4}```
[option 5]: ```{OPTION_5}```\end{verbatim}
\par\noindent\rule{0.5\textwidth}{0.4pt}
\normalsize 
    \label{tab:zs_prompt}

Where the input parameters are self-explanatory. It is worth mentioning that for BrainTeaser, where the multiple-choice options are four, the prompt is adjusted accordingly.

\paragraph{Few-shot Prompting Techniques}
\label{sec:fewshot}
In this approach, we provide a specific number of exemplars from each dataset’s training set before asking the model to solve a multiple-choice question from the test set. These exemplars, which include their correct answers, serve as in-context learning examples. By presenting these examples, we aim to guide the model and improve its performance on the subsequent unanswered tasks. This method leverages the provided context to enhance the model's ability to understand and solve new instances more effectively. The system prompt that is used here is the one of the zero-shot prompting with some minor changes. The number of examples that are used are 2, 4 and 8 as mentioned in section \ref{sec:method}.

Now, a key decision is how to select the exemplars that will be provided to the model before it attempts to answer the test set questions. The simplest and most straightforward approach is to randomly choose these answered exemplars from the training dataset. This method, referred to as \textbf{FS Rand} in our experiments, offers a baseline for comparison, as it doesn't apply any specific strategy for selecting the most relevant examples and relies purely on random sampling.

Another approach implemented in our experiments for selecting exemplars is based on semantic similarity. Specifically, we utilize the Sentence Transformers library \footnote{\href{https://www.sbert.net/}{SentenceTransformers}} \cite{reimers-2019-sentence-bert} in conjunction with the \textit{gte-large-en-v1.5} model \footnote{\href{https://huggingface.co/Alibaba-NLP/gte-large-en-v1.5}{Alibaba-NLP/gte-large-en-v1.5}} \cite{zhang2024mgtegeneralizedlongcontexttext}. This setup allows us to generate text embeddings and perform semantic similarity comparisons using cosine similarity between these embeddings.
For each experiment, we first generate text embeddings for all instances in the training datasets. Then, for each question in the test set, we generate its embedding and calculate its cosine similarity with all training set embeddings. Based on the number of in-context learning examples needed, we select the most similar training instances to use as exemplars. This method ensures that the selected examples are semantically relevant, aiming to improve the model's performance by presenting it with contextually aligned training examples. This approach is referred to as \textbf{FS Sim} in our experiments, as it focuses on semantic similarity to guide exemplar selection for improved task performance.

The system and user prompts used for both of the above settings are outlined as follows:
\par\noindent\rule{0.5\textwidth}{0.4pt}
\small \textbf{System Prompt:}\\
You will encounter a riddle that requires analytical thinking and reasoning to solve.\\
A riddle is a question or statement intentionally phrased so as to require ingenuity in ascertaining its answer or meaning, typically presented as a game.\\\

Different ideas can be used in these riddles:\\
    1. Riddles often employ misdirection, leading you away from the actual solution.\\
    2. They include elements with double meanings, requiring a keen eye for words with dual interpretations.\\
    3. Metaphorical wordplay adds another layer, urging you to decipher figurative language.\\
    4. Look out for exaggeration, as riddles may present overly dramatic details to divert your attention.\\
    5. Common phrases and sayings may hide within the puzzle, demanding familiarity.\\
    6. Associations and irony play a crucial role, introducing unexpected connections.\\
    7. Numerical puzzles can also be part of the mystery, requiring you to decode their significance.\\
    8. Elemental imagery, drawn from nature, might hold key descriptors.\\
    9. Rhyming and sound clues can add a poetic dimension.\\
    10. Word Puzzles: Pay attention to anagrams, acrostics, and other wordplay elements.\\
    11. Also, it is important to note you should decode the upcoming riddle using everyday logic and creativity.\\
    \\
Approach the riddle with everyday logic and creativity, avoiding supernatural explanations.\\

First, you'll encounter \textbf{X} examples with their answer provided similar to the riddle you will need to solve.\\
\\
Then you will be given an unsolved riddle and \textbf{X} options to choose the answer amongst them.\\
\\
\textbf{User Prompt:}
\begin{verbatim}
{EXAMPLES}\end{verbatim}
Answer the following riddle while taking into consideration the examples above. Choose the best and the most logical option from the available choices:
\begin{verbatim}
Riddle: ```
{RIDDLE}
```

Options:
[option 1]: ```{OPTION_1}```
[option 2]: ```{OPTION_2}```
[option 3]: ```{OPTION_3}```
[option 4]: ```{OPTION_4}```
[option 5]: ```{OPTION_5}```\end{verbatim}
\par\noindent\rule{0.5\textwidth}{0.4pt}
\normalsize 

Where the \textbf{EXAMPLES} refer to the few-shot examples selected for each test set instance, and the other input parameters are self-explanatory. It is worth mentioning that for BrainTeaser, where the multiple-choice options are four, the prompt is adjusted accordingly.

\paragraph{Chain of Thought Prompting Technique}
Another promising technique for enhancing model performance across various tasks is Chain of Thought (CoT) prompting. This approach involves not only providing answered examples but also including an explanation consisting of intermediate reasoning steps. By laying out the thought process step-by-step, CoT prompting significantly improves the capability of large language models to tackle complex reasoning tasks. In our task, the Chain of Thought (CoT) prompting technique is anticipated to outperform the previously mentioned methods. This expectation is supported by the performance improvements observed in the CommonsenseQA dataset \cite{talmor-etal-2019-commonsenseqa},  a domain similar to our setting that involves both lateral and vertical thinking. The dataset demonstrated significant improvements when using a manually composed set of two, four, or eight few-shot exemplars with CoT prompting, highlighting its effectiveness in eliciting successful reasoning.
In our case, these explanations were generated manually to ensure they align with human perceptions of the reasoning process required by the riddle. The explanation format we used follows the structure outlined by \cite{wei2023chainofthoughtpromptingelicitsreasoning} for few-shot exemplars in full chain-of-thought prompts for CommonsenseQA. Due to the similarity in the multiple-choice question-answer format of our datasets, it served as a good foundation for developing our own approach of creating explanations.
Below, we provide the system and user prompts used in the aforementioned technique:
\par\noindent\rule{0.5\textwidth}{0.4pt}
\small \textbf{System Prompt:}\\
You will encounter a riddle that requires analytical thinking and reasoning to solve.\\
A riddle is a question or statement intentionally phrased so as to require ingenuity in ascertaining its answer or meaning, typically presented as a game.\\\

Different ideas can be used in these riddles:\\
    1. Riddles often employ misdirection, leading you away from the actual solution.\\
    2. They include elements with double meanings, requiring a keen eye for words with dual interpretations.\\
    3. Metaphorical wordplay adds another layer, urging you to decipher figurative language.\\
    4. Look out for exaggeration, as riddles may present overly dramatic details to divert your attention.\\
    5. Common phrases and sayings may hide within the puzzle, demanding familiarity.\\
    6. Associations and irony play a crucial role, introducing unexpected connections.\\
    7. Numerical puzzles can also be part of the mystery, requiring you to decode their significance.\\
    8. Elemental imagery, drawn from nature, might hold key descriptors.\\
    9. Rhyming and sound clues can add a poetic dimension.\\
    10. Word Puzzles: Pay attention to anagrams, acrostics, and other wordplay elements.\\
    11. Also, it is important to note you should decode the upcoming riddle using everyday logic and creativity.\\
    \\
Approach the riddle with everyday logic and creativity, avoiding supernatural explanations.\\

First, you'll encounter \textbf{X} examples demonstrating analytical reasoning similar to the riddle you will need to solve.\\
\\
Then you will be given an unsolved riddle and five options to choose the answer amongst them. Let's think step by step and solve the riddle based on the examples provided above.\\
\\
\textbf{User Prompt:}
\begin{verbatim}
{EXAMPLES_COT}\end{verbatim}
Answer the following riddle while taking into consideration the examples above. Choose the best and the most logical option from the available choices:
\begin{verbatim}
Riddle: ```
{RIDDLE}
```

Options:
[option 1]: ```{OPTION_1}```
[option 2]: ```{OPTION_2}```
[option 3]: ```{OPTION_3}```
[option 4]: ```{OPTION_4}```
[option 5]: ```{OPTION_5}```\end{verbatim}
\par\noindent\rule{0.5\textwidth}{0.4pt}
\normalsize 

Where the \textbf{EXAMPLES\_COT} refer to the few-shot examples with manually generated explanations selected for each test set instance, and the other input parameters are self-explanatory. It is important to note that for BrainTeaser, where there are four multiple-choice options, the prompt is modified accordingly.

\section{Experimental Setup}
\label{sec:models}
In our experiments, we employed the Google Colab platform and Kaggle, leveraging various open-source Python packages such as Transformers, BitsAndBytes, Accelerate \cite{accelerate} and Sentence-Transformers. We also utilized the Amazon Bedrock API, which allowed us to access models without being constrained by hardware limitations, providing scalable and flexible model deployment. 

We specifically opted for instruction variations of the models because they aligned more closely with our task requirements. Instruction-tuned models generally offer better performance in scenarios where understanding and following specific instructions is crucial. This alignment ensures that the models are better equipped to handle the nuances of the tasks, leading to more effective and relevant outputs.

The temperature and repetition penalty values were determined through a series of exploratory experiments. To ensure consistency across our work, we systematically applied the same parameters whenever possible. For our experiments, we used a temperature of 0.5 and repetition penalties of either 1.0 or 1.15.

\paragraph{Llama 3}\cite{llama3modelcard} In our experiments, we chose two variations of the model: the 8B and 70B versions. The Llama3-8B\footnote{\href{https://huggingface.co/meta-llama/Meta-Llama-3-8B-Instruct}{meta-llama/Meta-Llama-3-8B-Instruct}} model was used for inference without quantization, which allowed it to deliver results with full precision. This approach was ideal when computational resources were sufficient and precision was critical.
On the other hand, the Llama3-70B \footnote{\href{https://huggingface.co/meta-llama/Meta-Llama-3-70B-Instruct}{meta-llama/Meta-Llama-3-70B-Instruct}}  model, due to hardware limitations, underwent quantization. This process reduced the model’s size and computational needs, making it accessible despite the constraints of our hardware. While quantization might lead to some loss of precision, it was necessary to deploy a model with more parameters, which otherwise would not have been feasible.

\paragraph{Mistral}\cite{jiang2023mistral7b}
In our experiments, we used the Mistral-7B-Instruct-v0.2\footnote{\href{https://huggingface.co/mistralai/Mistral-7B-Instruct-v0.2}{mistralai/Mistral-7B-Instruct-v0.2}} and the Mixtral-8x7B-Instruct-v0.1\footnote{\href{https://huggingface.co/mistralai/Mixtral-8x7B-Instruct-v0.1}{mistralai/Mixtral-8x7B-Instruct-v0.1}} in their unquantized forms. The Mistral 7B-Instruct v0.2 was selected for its strong instruction-following abilities, maintaining full precision. Similarly, the Mixtral-8x7B-Instruct-v0.1, which integrates eight 7B models, was used unquantized to benefit from its combined performance. Our goal was to explore the effectiveness of smaller models for our riddle tasks, and we preferred these instruction-tuned variations to ensure they were well-suited to our specific requirements, leading to more effective results.

\paragraph{Qwen2}\cite{yang2024qwen2technicalreport}
To further explore smaller model variations, we selected the Qwen2-7B-Instruct\footnote{\href{https://huggingface.co/Qwen/Qwen2-7B-Instruct}{Qwen/Qwen2-7B-Instruct}} version. The Qwen2 family of models, including this 7B-Instruct variant, is known for its strong performance in reasoning tasks. The Qwen2 models are designed with advanced instruction-following capabilities and are particularly effective at complex problem-solving and logical reasoning, making them well-suited for our objectives.

\section{Detailed Results}
\label{sec:full_results}
In this section, we present the detailed results for both datasets across all our experimental techniques. Due to the extensive nature of the experiments, the results for each dataset are organized into subtables. We begin with the BrainTeaser dataset, where the metrics are more detailed due to the dataset's structure, followed by the results for the RiddleSense dataset. The structure is organized by method per model, with results presented in descending order based on score for each method, rather than by the number of examples. This differs from the approach outlined in Section \ref{sec:results}.

\begin{table*}[h!]
    \centering
    \resizebox{\textwidth}{!}{%
    \begin{tabular}{lc|cccc|ccccc|c}
    \midrule
    \textbf{Model} & \textbf{Method} & \textbf{Num.Ex} & 
    \textbf{Task} & 
    \textbf{Temp} &  \textbf{Rep\_Pen} &
    \textbf{Original} & \textbf{Semantic} & \textbf{Context} & 
    \textbf{Ori. + Sem.} & \textbf{Ori. + Sem. + Con.} &
    \textbf{Average} \\
    \midrule
    \multicolumn{12}{c}{\textbf{Chain-of-Thought Zero-shot}} \\
    \midrule
  
    Meta-Llama-3-70B-Instruct &  CoT\_ZS (Q) & 0 & SP & 
    0.5 & 1.15 & 
    0.725 & 0.775 & 0.675 & 0.675 & 0.550 & 0.725 \\
    \cmidrule(lr){1-12}

    Mixtral-8x7B-Instruct-v0.1 &  CoT\_ZS & 0 & SP & 
    0.5 & 1.0 & 
    0.575 & 0.550 & 0.525 & 0.475 & 0.275 & 0.550 \\
    \cmidrule(lr){1-12}

    Meta-Llama-3-8B-Instruct &  CoT\_ZS & 0 & SP & 
    0.5 & 1.0 & 
    0.625 & 0.650 & 0.625 & 0.500 & 0.325 & 0.633 \\
    \cmidrule(lr){1-12}

    Mistral-7B-Instruct-v0.2 &  CoT\_ZS & 0 & SP & 
    0.5 & 1.0 & 
    0.375 & 0.475 & 0.500 & 0.300 & 0.250 & 0.450 \\
    \cmidrule(lr){1-12}

    Qwen2-7B-Instruct &  CoT\_ZS & 0 & SP & 
    0.5 & 1.15 & 
    0.475 & 0.450 & 0.450 & 0.350 & 0.200 & 0.458 \\

    \midrule
    \multicolumn{12}{c}{\textbf{Few-shot with CoT Explanations }} \\
    \midrule
    
    Meta-Llama-3-70B-Instruct &  CoT\_FS (Q) & 2 & SP & 
    0.5 & 1.15 &
    0.850 & 0.750 & 0.675 & 0.700 & 0.500 & 0.758 \\
    \cmidrule(lr){2-12}

    Meta-Llama-3-70B-Instruct &  CoT\_FS (Q) & 8 & SP & 
    0.5 & 1.0 & 
    0.675 & 0.750 & 0.700 & 0.600 & 0.425 & 0.708 \\
    \cmidrule(lr){2-12}

    Meta-Llama-3-70B-Instruct &  CoT\_FS (Q) & 4 & SP & 
    0.5 & 1.15 &
    0.700 & 0.650 & 0.700 & 0.575 & 0.450 & 0.683 \\
    \cmidrule(lr){1-12}

    Mixtral-8x7B-v0.1 &  CoT\_FS & 8 & SP & 
    0.5 & 1.0 & 
    0.650 & 0.675 & 0.600 & 0.575 & 0.375 & 0.642 \\
    \cmidrule(lr){2-12}

    Mixtral-8x7B-v0.1 &  CoT\_FS & 2 & SP & 
    0.5 & 1.0 & 
    0.625 & 0.625 & 0.600 & 0.525 & 0.350 & 0.617 \\
    \cmidrule(lr){2-12}

    Mixtral-8x7B-v0.1 &  CoT\_FS & 4 & SP & 
    0.5 & 1.0 & 
    0.575 & 0.600 & 0.575 & 0.475 & 0.350 & 0.583 \\
    \cmidrule(lr){1-12}
    
    Meta-Llama-3-8B-Instruct &  CoT\_FS & 8 & SP & 
    0.5 & 1.0 & 
    0.675 & 0.700 & 0.600 & 0.525 & 0.325 & 0.658 \\
    \cmidrule(lr){2-12}

    Meta-Llama-3-8B-Instruct &  CoT\_FS & 2 & SP & 
    0.5 & 1.0 & 
    0.725 & 0.625 & 0.550 & 0.550 & 0.350 & 0.633 \\
    \cmidrule(lr){2-12}

    Meta-Llama-3-8B-Instruct &  CoT\_FS & 4 & SP & 
    0.5 & 1.0 & 
    0.650 & 0.625 & 0.550 & 0.500 & 0.300 & 0.608 \\
    \cmidrule(lr){1-12}

    Mistral-7B-Instruct-v0.2 &  CoT\_FS & 8 & SP & 
    0.5 & 1.0 & 
    0.525 & 0.550 & 0.450 & 0.375 & 0.300 & 0.508 \\
    \cmidrule(lr){2-12}

    Mistral-7B-Instruct-v0.2 &  CoT\_FS & 4 & SP & 
    0.5 & 1.0 & 
    0.525 & 0.550 & 0.450 & 0.375 & 0.300 & 0.508 \\
    \cmidrule(lr){2-12}

    Mistral-7B-Instruct-v0.2 &  CoT\_FS & 2 & SP & 
    0.5 & 1.0 & 
    0.525 & 0.425 & 0.475 & 0.300 & 0.225 & 0.475 \\
    \cmidrule(lr){1-12}

    Qwen2-7B-Instruct &  CoT\_FS & 8 & SP & 
    0.5 & 1.0 & 
    0.600 & 0.725 & 0.675 & 0.550 & 0.425 & 0.667 \\
    \cmidrule(lr){2-12}

    Qwen2-7B-Instruct &  CoT\_FS & 4 & SP & 
    0.5 & 1.0 & 
    0.650 & 0.675 & 0.625 & 0.550 & 0.450 & 0.650 \\
    \cmidrule(lr){2-12}

    Qwen2-7B-Instruct &  CoT\_FS & 2 & SP & 
    0.5 & 1.0 & 
    0.675 & 0.550 & 0.600 & 0.450 & 0.375 & 0.608 \\
 
    \midrule
    \multicolumn{12}{c}{\textbf{Few-shot with Random Selection}} \\
    \midrule
    
    Meta-Llama-3-70B-Instruct & FS Rand (Q) & 4 & SP & 
    0.5 & 1.15 & 
    0.825 & 0.850 & 0.750 & 0.800 & 0.700 & 0.808 \\
    \cmidrule(lr){2-12}

    Meta-Llama-3-70B-Instruct & FS Rand (Q) & 8 & SP & 
    0.5 & 1.15 & 
    0.775 & 0.800 & 0.750 & 0.675 & 0.550 & 0.775 \\
    \cmidrule(lr){2-12}

    Meta-Llama-3-70B-Instruct & FS Rand (Q) & 2 & SP & 
    0.5 & 1.15 & 
    0.750 & 0.800 & 0.775 & 0.725 & 0.600 & 0.775 \\
    \cmidrule(lr){1-12}

    Mixtral-8x7B-Instruct-v0.1 & FS Rand & 4 & SP & 
    0.5 & 1.0 & 
    0.675 & 0.775 & 0.600 & 0.600 & 0.400 & 0.683 \\
    \cmidrule(lr){2-12}

    Mixtral-8x7B-Instruct-v0.1 & FS Rand & 2 & SP & 
    0.5 & 1.0 & 
    0.650 & 0.625 & 0.575 & 0.475 & 0.300 & 0.617 \\
    \cmidrule(lr){2-12}

    Mixtral-8x7B-Instruct-v0.1 & FS Rand & 8 & SP & 
    0.5 & 1.0 & 
    0.675 & 0.650 & 0.525 & 0.525 & 0.350 & 0.617 \\
    \cmidrule(lr){1-12}

    Meta-Llama-3-8B-Instruct & FS Rand & 8 & SP & 
    0.5 & 1.0 & 
    0.750 & 0.625 & 0.650 & 0.575 & 0.400 & 0.675 \\
    \cmidrule(lr){2-12}

    Meta-Llama-3-8B-Instruct & FS Rand & 4 & SP & 
    0.5 & 1.0 & 
    0.625 & 0.650 & 0.650 & 0.500 & 0.375 & 0.642 \\
    \cmidrule(lr){2-12}

    Meta-Llama-3-8B-Instruct & FS Rand & 2 & SP & 
    0.5 & 1.0 & 
    0.600 & 0.700 & 0.600 & 0.550 & 0.375 & 0.633 \\
    \cmidrule(lr){1-12}

    Mistral-7B-Instruct-v0.2 & FS Rand & 2 & SP & 
    0.5 & 1.0 & 
    0.525 & 0.550 & 0.475 & 0.425 & 0.275 & 0.517 \\
    \cmidrule(lr){2-12}

    Mistral-7B-Instruct-v0.2 & FS Rand & 8 & SP & 
    0.5 & 1.0 & 
    0.450 & 0.575 & 0.425 & 0.400 & 0.300 & 0.483 \\
    \cmidrule(lr){2-12}

    Mistral-7B-Instruct-v0.2 & FS Rand & 4 & SP & 
    0.5 & 1.0 & 
    0.550 & 0.425 & 0.475 & 0.325 & 0.225 & 0.483 \\
    \cmidrule(lr){1-12}

    Qwen2-7B-Instruct & FS Rand & 2 & SP & 
    0.5 & 1.15 & 
    0.675 & 0.650 & 0.600 & 0.550 & 0.425 & 0.642 \\
    \cmidrule(lr){2-12}

    Qwen2-7B-Instruct & FS Rand & 8 & SP & 
    0.5 & 1.15 & 
    0.700 & 0.600 & 0.625 & 0.525 & 0.425 & 0.642 \\
    \cmidrule(lr){2-12}

    Qwen2-7B-Instruct & FS Rand & 4 & SP & 
    0.5 & 1.15 & 
    0.675 & 0.575 & 0.575 & 0.525 & 0.400 & 0.608 \\

    \midrule
    \end{tabular}%
    }
\caption{Model performance for \textit{BrainTeaser} (Part 1). The presence of \textbf{(Q)} in the method column indicates that the results correspond to the quantized version of the model.}
\label{tab:sentence_results1}
\end{table*}
\begin{table*}[h!]
    \centering
    \resizebox{\textwidth}{!}{%
    \begin{tabular}{lc|cccc|ccccc|c}
    \midrule
    \textbf{Model} & \textbf{Method} & \textbf{Num.Ex} & 
    \textbf{Task} & 
    \textbf{Temp} &  \textbf{Rep\_Pen} &
    \textbf{Original} & \textbf{Semantic} & \textbf{Context} & 
    \textbf{Ori. + Sem.} & \textbf{Ori. + Sem. + Con.} &
    \textbf{Average} \\

    \midrule
    \multicolumn{12}{c}{\textbf{Few-shot with Semantic Similarity}} \\
    \midrule

    Meta-Llama-3-70B-Instruct & FS Sim (Q) & 2 & SP & 
    0.5 & 1.15 & 
    0.850 & 0.875 & 0.750 & 0.800 & 0.675 & 0.825 \\
    \cmidrule(lr){2-12}
    
    Meta-Llama-3-70B-Instruct & FS Sim (Q) & 4 & SP & 
    0.5 & 1.15 & 
    0.825 & 0.775 & 0.775 & 0.700 & 0.575 & 0.792 \\
    \cmidrule(lr){2-12}
    
    Meta-Llama-3-70B-Instruct & FS Sim (Q) & 8 & SP & 
    0.5 & 1.15 & 
    0.775 & 0.825 & 0.750 & 0.700 & 0.600 & 0.783 \\
    \cmidrule(lr){1-12}

    Mixtral-8x7B-Instruct-v0.1 & FS Sim & 2 & SP & 
    0.5 & 1.0 & 
    0.700 & 0.700 & 0.675 & 0.575 & 0.450 & 0.692 \\
    \cmidrule(lr){2-12}

    Mixtral-8x7B-Instruct-v0.1 & FS Sim & 4 & SP & 
    0.5 & 1.0 & 
    0.750 & 0.650 & 0.650 & 0.575 & 0.425 & 0.683 \\
    \cmidrule(lr){2-12}
    
    Mixtral-8x7B-Instruct-v0.1 & FS Sim & 8 & SP & 
    0.5 & 1.0 & 
    0.750 & 0.625 & 0.625 & 0.525 & 0.375 & 0.667  \\
    \cmidrule(lr){1-12}

    Meta-Llama-3-8B-Instruct & FS Sim & 8 & SP & 
    0.5 & 1.0 & 
    0.750 & 0.775 & 0.775 & 0.700 & 0.600 & 0.767 \\
    \cmidrule(lr){2-12}

    Meta-Llama-3-8B-Instruct & FS Sim & 4 & SP & 
    0.5 & 1.0 & 
    0.675 & 0.825 & 0.650 & 0.675 & 0.475 & 0.717 \\
    \cmidrule(lr){2-12}

    Meta-Llama-3-8B-Instruct & FS Sim & 2 & SP & 
    0.5 & 1.0 & 
    0.800 & 0.675 & 0.625 & 0.625 & 0.525 & 0.700 \\
    \cmidrule(lr){1-12}

    Mistral-7B-Instruct-v0.2 & FS Sim & 8 & SP & 
    0.5 & 1.0 & 
    0.550 & 0.600 & 0.450 & 0.500 & 0.325 & 0.533 \\
    \cmidrule(lr){2-12}
    
    Mistral-7B-Instruct-v0.2 & FS Sim & 2 & SP & 
    0.5 & 1.0 & 
    0.55 & 0.475 & 0.525 & 0.375 & 0.275 & 0.517 \\
    \cmidrule(lr){2-12}

    Mistral-7B-Instruct-v0.2 & FS Sim & 4 & SP & 
    0.5 & 1.0 & 
    0.475 & 0.450 & 0.450 & 0.350 & 0.225 & 0.458 \\
    \cmidrule(lr){1-12}

    Qwen2-7B-Instruct & FS Sim & 8 & SP & 
    0.5 & 1.15 & 
    0.675 & 0.625 & 0.650 & 0.575 & 0.500 & 0.650 \\
    \cmidrule(lr){2-12}

    Qwen2-7B-Instruct & FS Sim & 4 & SP & 
    0.5 & 1.15 & 
    0.650 & 0.600 & 0.650 & 0.550 & 0.450 & 0.633 \\
    \cmidrule(lr){2-12}

    Qwen2-7B-Instruct & FS Sim & 2 & SP & 
    0.5 & 1.15 & 
    0.600 & 0.550 & 0.650 & 0.500 & 0.400 & 0.600 \\

    \midrule
    \multicolumn{12}{c}{\textbf{RISCORE\textsubscript{m} Rand}} \\
    \midrule

    Meta-Llama-3-70B-Instruct & RISCORE\textsubscript{m} Rand (Q) & 8 & SP & 
    0.5 & 1.15 & 
    0.775 & 0.825 & 0.800 & 0.700 & 0.600 & 0.800 \\
    \cmidrule(lr){2-12}
    
    Meta-Llama-3-70B-Instruct & RISCORE\textsubscript{m} Rand (Q) & 2 & SP & 
    0.5 & 1.15 & 
    0.850 & 0.775 & 0.725 & 0.725 & 0.650 & 0.783 \\
    \cmidrule(lr){2-12}

    Meta-Llama-3-70B-Instruct & RISCORE\textsubscript{m} Rand (Q) & 4 & SP & 
    0.5 & 1.15 & 
    0.775 & 0.800 & 0.700 & 0.725 & 0.600 & 0.758 \\
    \cmidrule(lr){1-12}

    Mixtral-8x7B-Instruct-v0.1 & RISCORE\textsubscript{m} Rand & 8 & SP & 
    0.5 & 1.0 & 
    0.700 & 0.625 & 0.625 & 0.600 & 0.450 & 0.650 \\
    \cmidrule(lr){2-12}

    Mixtral-8x7B-Instruct-v0.1 & RISCORE\textsubscript{m} Rand & 2 & SP & 
    0.5 & 1.0 & 
    0.725 & 0.600 & 0.550 & 0.600 & 0.400 & 0.625 \\
    \cmidrule(lr){2-12}

    Mixtral-8x7B-Instruct-v0.1 & RISCORE\textsubscript{m} Rand & 4 & SP & 
    0.5 & 1.0 & 
    0.675 & 0.650 & 0.525 & 0.550 & 0.325 & 0.617 \\
    \cmidrule(lr){1-12}

    Meta-Llama-3-8B-Instruct & RISCORE\textsubscript{m} Rand & 4 & SP & 
    0.5 & 1.0 & 
    0.650 & 0.700 & 0.675 & 0.625 & 0.500 & 0.675 \\
    \cmidrule(lr){2-12}
    
    Meta-Llama-3-8B-Instruct & RISCORE\textsubscript{m} Rand & 2 & SP & 
    0.5 & 1.0 & 
    0.700 & 0.725 & 0.575 & 0.525 & 0.375 & 0.667 \\
    \cmidrule(lr){2-12}

    Meta-Llama-3-8B-Instruct & RISCORE\textsubscript{m} Rand & 8 & SP & 
    0.5 & 1.0 & 
    0.700 & 0.750 & 0.550 & 0.575 & 0.425 & 0.667 \\
    \cmidrule(lr){1-12}
    
    Mistral-7B-Instruct-v0.2 & RISCORE\textsubscript{m} Rand & 4 & SP & 
    0.5 & 1.0 & 
    0.575 & 0.475 & 0.500 & 0.425 & 0.300 & 0.517 \\
    \cmidrule(lr){2-12}

    Mistral-7B-Instruct-v0.2 & RISCORE\textsubscript{m} Rand & 2 & SP & 
    0.5 & 1.0 & 
    0.550 & 0.425 & 0.400 & 0.325 & 0.200 & 0.458 \\
    \cmidrule(lr){2-12}
    
    Mistral-7B-Instruct-v0.2 & RISCORE\textsubscript{m} Rand & 8 & SP & 
    0.5 & 1.0 & 
    0.350 & 0.450 & 0.400 & 0.275 & 0.200 & 0.400 \\
    \cmidrule(lr){1-12}

    Qwen2-7B-Instruct & RISCORE\textsubscript{m} Rand & 4 & SP & 
    0.5 & 1.15 & 
    0.625 & 0.650 & 0.600 & 0.550 & 0.400 & 0.625 \\
    \cmidrule(lr){2-12}

    Qwen2-7B-Instruct & RISCORE\textsubscript{m} Rand & 2 & SP & 
    0.5 & 1.15 & 
    0.650 & 0.575 & 0.600 & 0.500 & 0.400 & 0.608 \\
    \cmidrule(lr){2-12}

    Qwen2-7B-Instruct & RISCORE\textsubscript{m} Rand & 8 & SP & 
    0.5 & 1.15 & 
    0.625 & 0.550 & 0.600 & 0.450 & 0.325 & 0.592 \\

    \midrule
    \multicolumn{12}{c}{\textbf{RISCORE\textsubscript{m} Sim}} \\
    \midrule
    
    Meta-Llama-3-70B-Instruct & RISCORE\textsubscript{m} Sim (Q) & 4 & SP & 
    0.5 & 1.15 & 
    0.850 & 0.850 & 0.800 & 0.800 & 0.650 & 0.833 \\
    \cmidrule(lr){2-12}

    Meta-Llama-3-70B-Instruct & RISCORE\textsubscript{m} Sim (Q) & 8 & SP & 
    0.5 & 1.15 & 
    0.800 & 0.775 & 0.850 & 0.700 & 0.625 & 0.808 \\
    \cmidrule(lr){2-12}

    Meta-Llama-3-70B-Instruct & RISCORE\textsubscript{m} Sim (Q) & 2 & SP & 
    0.5 & 1.15 & 
    0.850 & 0.775 & 0.725 & 0.725 & 0.600 & 0.783 \\
    \cmidrule(lr){1-12}

    Mixtral-8x7B-Instruct-v0.1 & RISCORE\textsubscript{m} Sim & 8 & SP & 
    0.5 & 1.0 & 
    0.775 & 0.625 & 0.725 & 0.575 & 0.500 & 0.708 \\
    \cmidrule(lr){2-12}
    
    Mixtral-8x7B-Instruct-v0.1 & RISCORE\textsubscript{m} Sim & 4 & SP & 
    0.5 & 1.0 & 
    0.675 & 0.750 & 0.700 & 0.600 & 0.425 & 0.708 \\
    \cmidrule(lr){2-12}

    Mixtral-8x7B-Instruct-v0.1 & RISCORE\textsubscript{m} Sim & 2 & SP & 
    0.5 & 1.0 & 
    0.725 & 0.675 & 0.625 & 0.675 & 0.475 & 0.675 \\
    \cmidrule(lr){1-12}

    Meta-Llama-3-8B-Instruct & RISCORE\textsubscript{m} Sim & 2 & SP & 
    0.5 & 1.0 & 
    0.750 & 0.875 & 0.675 & 0.675 & 0.575 & 0.767 \\
    \cmidrule(lr){2-12}

    Meta-Llama-3-8B-Instruct & RISCORE\textsubscript{m} Sim & 8 & SP & 
    0.5 & 1.0 & 
    0.775 & 0.800 & 0.700 & 0.775 & 0.650 & 0.758 \\
    \cmidrule(lr){2-12}

    Meta-Llama-3-8B-Instruct & RISCORE\textsubscript{m} Sim & 4 & SP & 
    0.5 & 1.0 & 
    0.825 & 0.700 & 0.700 & 0.675 & 0.525 & 0.742 \\
    \cmidrule(lr){1-12}

    Mistral-7B-Instruct-v0.2 & RISCORE\textsubscript{m} Sim & 4 & SP & 
    0.5 & 1.0 & 
    0.575 & 0.550 & 0.575 & 0.400 & 0.325 & 0.567 \\
    \cmidrule(lr){2-12}

    Mistral-7B-Instruct-v0.2 & RISCORE\textsubscript{m} Sim & 8 & SP & 
    0.5 & 1.0 & 
    0.575 & 0.575 & 0.500 & 0.450 & 0.325 & 0.550 \\
    \cmidrule(lr){2-12}

    Mistral-7B-Instruct-v0.2 & RISCORE\textsubscript{m} Sim & 2 & SP & 
    0.5 & 1.0 & 
    0.500 & 0.475 & 0.475 & 0.350 & 0.200 & 0.483 \\

    \cmidrule(lr){1-12}

    Qwen2-7B-Instruct & RISCORE\textsubscript{m} Sim & 8 & SP & 
    0.5 & 1.15 & 
    0.700 & 0.650 & 0.650 & 0.575 & 0.425 & 0.667 \\
    \cmidrule(lr){2-12}

    Qwen2-7B-Instruct & RISCORE\textsubscript{m} Sim & 2 & SP & 
    0.5 & 1.15 & 
    0.700 & 0.625 & 0.675 & 0.550 & 0.375 & 0.667 \\
    \cmidrule(lr){2-12}

    Qwen2-7B-Instruct & RISCORE\textsubscript{m} Sim & 4 & SP & 
    0.5 & 1.15 & 
    0.750 & 0.625 & 0.550 & 0.575 & 0.425 & 0.642 \\

    \midrule
    \end{tabular}%
    }
\caption{Model performance for \textit{BrainTeaser} (Part 2). The presence of \textbf{(Q)} in the method column indicates that the results correspond to the quantized version of the model.}
\label{tab:sentence_results2}
\end{table*}
\begin{table*}[h!]
    \centering
    \resizebox{\textwidth}{!}{%
    \begin{tabular}{lc|cccc|ccccc|c}
    \midrule
    \textbf{Model} & \textbf{Method} & \textbf{Num.Ex} & 
    \textbf{Task} & 
    \textbf{Temp} &  \textbf{Rep\_Pen} &
    \textbf{Original} & \textbf{Semantic} & \textbf{Context} & 
    \textbf{Ori. + Sem.} & \textbf{Ori. + Sem. + Con.} &
    \textbf{Average} \\

 \midrule
    \multicolumn{12}{c}{\textbf{RISCORE Results}} \\
    \midrule
    \multicolumn{12}{c}{Llama3-70B zeroshot for QA \& Llama3-8B for distractors} \\
    \midrule

    Meta-Llama-3-70B-Instruct & RISCORE Sim (Q) & 8 & SP & 
    0.5 & 1.15 & 
    0.850 & 0.825 & 0.750 & 0.775 & 0.600 & 0.808 \\
    \cmidrule(lr){2-12}
    
    Meta-Llama-3-70B-Instruct & RISCORE Sim (Q) & 2 & SP & 
    0.5 & 1.15 & 
    0.800 & 0.775 & 0.800 & 0.750 & 0.650 & 0.792 \\
    \cmidrule(lr){2-12}

    Meta-Llama-3-70B-Instruct & RISCORE Sim (Q) & 4 & SP & 
    0.5 & 1.15 & 
    0.825 & 0.750 & 0.800 & 0.675 & 0.600 & 0.792 \\
    \cmidrule(lr){1-12}

    Mixtral-8x7B-Instruct-v0.1 & RISCORE Sim & 8 & SP & 
    0.5 & 1.0 & 
    0.700 & 0.750 & 0.600 & 0.650 & 0.475 & 0.683 \\
    \cmidrule(lr){2-12}

    Mixtral-8x7B-Instruct-v0.1 & RISCORE Sim & 2 & SP & 
    0.5 & 1.0 & 
    0.750 & 0.750 & 0.500 & 0.650 & 0.375 & 0.667 \\
    \cmidrule(lr){2-12}

    Mixtral-8x7B-Instruct-v0.1 & RISCORE Sim & 4 & SP & 
    0.5 & 1.0 & 
    0.725 & 0.625 & 0.575 & 0.575 & 0.375 & 0.642 \\
    \cmidrule(lr){1-12}

    Meta-Llama-3-8B-Instruct & RISCORE Sim & 8 & SP & 
    0.5 & 1.0 & 
    0.700 & 0.725 & 0.675 & 0.650 & 0.525 & 0.700 \\
    \cmidrule(lr){2-12}

    Meta-Llama-3-8B-Instruct & RISCORE Sim & 4 & SP & 
    0.5 & 1.0 & 
    0.650 & 0.675 & 0.700 & 0.550 & 0.425 & 0.675 \\
    \cmidrule(lr){2-12}

    Meta-Llama-3-8B-Instruct & RISCORE Sim & 2 & SP & 
    0.5 & 1.0 & 
    0.700 & 0.600 & 0.575 & 0.525 & 0.350 & 0.625 \\
    \cmidrule(lr){1-12}

    Mistral-7B-Instruct-v0.2 & RISCORE Sim & 2 & SP & 
    0.5 & 1.0 & 
    0.550 & 0.450 & 0.475 & 0.375 & 0.300 & 0.492 \\
    \cmidrule(lr){2-12}

    Mistral-7B-Instruct-v0.2 & RISCORE Sim & 4 & SP & 
    0.5 & 1.0 & 
    0.525 & 0.400 & 0.475 & 0.300 & 0.200 & 0.467 \\
    \cmidrule(lr){2-12}

    Mistral-7B-Instruct-v0.2 & RISCORE Sim & 8 & SP & 
    0.5 & 1.0 & 
    0.425 & 0.525 & 0.475 & 0.400 & 0.325 & 0.475 \\
    \cmidrule(lr){1-12}

    Qwen2-7B-Instruct & RISCORE Sim & 8 & SP & 
    0.5 & 1.15 & 
    0.600 & 0.700 & 0.625 & 0.550 & 0.425 & 0.642 \\
    \cmidrule(lr){2-12}
    
    Qwen2-7B-Instruct & RISCORE Sim & 2 & SP & 
    0.5 & 1.15 & 
    0.650 & 0.650 & 0.575 & 0.550 & 0.400 & 0.625 \\
    \cmidrule(lr){2-12}

    Qwen2-7B-Instruct & RISCORE Sim & 4 & SP & 
    0.5 & 1.15 & 
    0.600 & 0.650 & 0.625 & 0.550 & 0.450 & 0.625 \\

    \midrule
    \multicolumn{12}{c}{Llama3-70B fewshot for QA \& Llama3-8B for distractors} \\
    \midrule
    
    Meta-Llama-3-70B-Instruct & RISCORE Sim (Q) & 8 & SP & 
    0.5 & 1.15 & 
    0.800 & 0.800 & 0.825 & 0.750 & 0.650 & 0.808 \\
    \cmidrule(lr){2-12}

    Meta-Llama-3-70B-Instruct & RISCORE Sim (Q) & 4 & SP & 
    0.5 & 1.15 & 
    0.775 & 0.800 & 0.800 & 0.700 & 0.600 & 0.792 \\
    \cmidrule(lr){2-12}

    Meta-Llama-3-70B-Instruct & RISCORE Sim (Q) & 2 & SP & 
    0.5 & 1.15 & 
    0.800 & 0.725 & 0.725 & 0.725 & 0.600 & 0.750 \\
    \cmidrule(lr){1-12}

    Mixtral-8x7B-Instruct-v0.1 & RISCORE Sim & 2 & SP & 
    0.5 & 1.0 & 
    0.725 & 0.725 & 0.575 & 0.675 & 0.475 & 0.675 \\
    \cmidrule(lr){2-12}

    Mixtral-8x7B-Instruct-v0.1 & RISCORE Sim & 8 & SP & 
    0.5 & 1.0 & 
    0.700 & 0.675 & 0.65 & 0.625 & 0.450 & 0.675 \\
    \cmidrule(lr){2-12}

    Mixtral-8x7B-Instruct-v0.1 & RISCORE Sim & 4 & SP & 
    0.5 & 1.0 & 
    0.700 & 0.675 & 0.575 & 0.625 & 0.400 & 0.650 \\
    \cmidrule(lr){1-12}

    Meta-Llama-3-8B-Instruct & RISCORE Sim & 8 & SP & 
    0.5 & 1.0 & 
    0.775 & 0.750 & 0.700 & 0.700 & 0.600 & 0.742 \\
    \cmidrule(lr){2-12}

    Meta-Llama-3-8B-Instruct & RISCORE Sim & 2 & SP & 
    0.5 & 1.0 & 
    0.725 & 0.700 & 0.625 & 0.600 & 0.400 & 0.683 \\
    \cmidrule(lr){2-12}

    Meta-Llama-3-8B-Instruct & RISCORE Sim & 4 & SP & 
    0.5 & 1.0 & 
    0.725 & 0.625 & 0.625 & 0.550 & 0.450 & 0.658 \\
    \cmidrule(lr){1-12}

    Mistral-7B-Instruct-v0.2 & RISCORE Sim & 4 & SP & 
    0.5 & 1.0 & 
    0.625 & 0.575 & 0.475 & 0.475 & 0.350 & 0.558 \\
    \cmidrule(lr){2-12}

    Mistral-7B-Instruct-v0.2 & RISCORE Sim & 8 & SP & 
    0.5 & 1.0 & 
    0.500 & 0.550 & 0.500 & 0.425 & 0.350 & 0.517 \\
    \cmidrule(lr){2-12}

    Mistral-7B-Instruct-v0.2 & RISCORE Sim & 2 & SP & 
    0.5 & 1.0 & 
    0.550 & 0.425 & 0.450 & 0.375 & 0.225 & 0.475 \\
    \cmidrule(lr){1-12}
    
    Qwen2-7B-Instruct & RISCORE Sim & 4 & SP & 
    0.5 & 1.15 & 
    0.625 & 0.700 & 0.650 & 0.575 & 0.450 & 0.658 \\
    \cmidrule(lr){2-12}

    Qwen2-7B-Instruct & RISCORE Sim & 8 & SP & 
    0.5 & 1.15 & 
    0.650 & 0.675 & 0.650 & 0.575 & 0.475 & 0.658 \\
    \cmidrule(lr){2-12}

    Qwen2-7B-Instruct & RISCORE Sim & 2 & SP & 
    0.5 & 1.15 & 
    0.625 & 0.650 & 0.600 & 0.525 & 0.400 & 0.625 \\
    
    \midrule
    \multicolumn{12}{c}{Llama3-70B fewshot for QA \& Llama3-70B for distractors} \\
    \midrule

    Meta-Llama-3-70B-Instruct & RISCORE Sim (Q) & 4 & SP & 
    0.5 & 1.15 & 
    0.875 & 0.775 & 0.725 & 0.750 & 0.600 & 0.792 \\
    \cmidrule(lr){2-12}

    Meta-Llama-3-70B-Instruct & RISCORE Sim (Q) & 2 & SP & 
    0.5 & 1.15 & 
    0.775 & 0.825 & 0.750 & 0.775 & 0.675 & 0.783 \\
    \cmidrule(lr){2-12}

    Meta-Llama-3-70B-Instruct & RISCORE Sim (Q) & 8 & SP & 
    0.5 & 1.15 & 
    0.775 & 0.750 & 0.775 & 0.725 & 0.600 & 0.767 \\
    \cmidrule(lr){1-12}

    Mixtral-8x7B-Instruct-v0.1 & RISCORE Sim & 8 & SP & 
    0.5 & 1.0 & 
    0.700 & 0.725 & 0.625 & 0.600 & 0.450 & 0.683 \\
    \cmidrule(lr){2-12}

    Mixtral-8x7B-Instruct-v0.1 & RISCORE Sim & 2 & SP & 
    0.5 & 1.0 & 
    0.700 & 0.700 & 0.600 & 0.625 & 0.500 & 0.667 \\
    \cmidrule(lr){2-12}

    Mixtral-8x7B-Instruct-v0.1 & RISCORE Simm & 4 & SP & 
    0.5 & 1.0 & 
    0.725 & 0.650 & 0.550 & 0.600 & 0.425 & 0.642 \\
    \cmidrule(lr){1-12}
    
    Meta-Llama-3-8B-Instruct & RISCORE Sim & 8 & SP & 
    0.5 & 1.0 & 
    0.800 & 0.675 & 0.625 & 0.625 & 0.475 & 0.700 \\
    \cmidrule(lr){2-12}

    Meta-Llama-3-8B-Instruct & RISCORE Sim & 2 & SP & 
    0.5 & 1.0 & 
    0.675 & 0.700 & 0.675 & 0.600 & 0.475 & 0.683 \\
    \cmidrule(lr){2-12}

    Meta-Llama-3-8B-Instruct & RISCORE Sim & 4 & SP & 
    0.5 & 1.0 & 
    0.725 & 0.650 & 0.625 & 0.550 & 0.475 & 0.667 \\
    \cmidrule(lr){1-12}

    Mistral-7B-Instruct-v0.2 & RISCORE Sim & 4 & SP & 
    0.5 & 1.0 & 
    0.575 & 0.500 & 0.450 & 0.375 & 0.300 & 0.508 \\
    \cmidrule(lr){2-12}

    Mistral-7B-Instruct-v0.2 & RISCORE Sim & 2 & SP & 
    0.5 & 1.0 & 
    0.625 & 0.400 & 0.475 & 0.350 & 0.275 & 0.500 \\
    \cmidrule(lr){2-12}

    Mistral-7B-Instruct-v0.2 & RISCORE Sim & 8 & SP & 
    0.5 & 1.0 & 
    0.600 & 0.475 & 0.425 & 0.400 & 0.300 & 0.500 \\
    \cmidrule(lr){1-12}

    Qwen2-7B-Instruct & RISCORE Sim & 2 & SP & 
    0.5 & 1.15 & 
    0.650 & 0.600 & 0.600 & 0.500 & 0.400 & 0.617 \\
    \cmidrule(lr){2-12}

    Qwen2-7B-Instruct & RISCORE Sim & 4 & SP & 
    0.5 & 1.15 & 
    0.650 & 0.625 & 0.575 & 0.600 & 0.475 & 0.617 \\
    \cmidrule(lr){2-12}

    Qwen2-7B-Instruct & RISCORE Sim & 8 & SP & 
    0.5 & 1.15 & 
    0.625 & 0.625 & 0.600 & 0.575 & 0.450 & 0.617 \\

    \midrule
    \end{tabular}%
    }
\caption{Model performance for \textit{BrainTeaser} (Part 3). The presence of \textbf{(Q)} in the method column indicates that the results correspond to the quantized version of the model.}
\label{tab:sentence_results3}
\end{table*}

\begin{table*}[h!]
    \centering
    \small
    \resizebox{!}{0.5\textheight}{%
    \begin{tabular}{lc|cccc|c}
    \midrule
    \textbf{Model} & \textbf{Method} & \textbf{Num.Ex} & 
    \textbf{Quant} & 
    \textbf{Temp} &  \textbf{Rep\_Pen} &
    \textbf{Average} \\
    \midrule
    \multicolumn{7}{c}{\textbf{Chain-of-Thought Zero-shot}} \\
    \midrule

    Meta-Llama-3-70B-Instruct & CoT\_ZS (Q) & 0 & 4bit & 
    0.5 & 1.15 & 
    0.775 \\
    \cmidrule(lr){1-7}

    Mixtral-8x7B-v0.1 & CoT\_ZS & 0 & False & 
    0.5 & 1.0 & 
    0.675 \\
    \cmidrule(lr){1-7}

    Meta-Llama-3-8B-Instruct & CoT\_ZS & 0 & False & 
    0.5 & 1.0 & 
    0.619 \\
    \cmidrule(lr){1-7}

    Mistral-7B-Instruct-v0.2 & CoT\_ZS & 0 & False & 
    0.5 & 1.0 & 
    0.589 \\
    \cmidrule(lr){1-7}

    Qwen2-7B-Instruct & CoT\_ZS & 0 & False & 0.5 & 1.15 & 0.608 \\
    \cmidrule(lr){1-7}


    \midrule
    \multicolumn{7}{c}{\textbf{Few-shot with CoT Explanations}} \\
    \midrule

    Meta-Llama-3-70B-Instruct & CoT\_FS (Q) & 2 & 4bit & 
    0.5 & 1.15 & 
    0.789 \\
    \cmidrule(lr){2-7}

    Meta-Llama-3-70B-Instruct & CoT\_FS (Q) & 4 & 4bit & 
    0.5 & 1.15 & 
    0.783 \\
    \cmidrule(lr){2-7}

    Meta-Llama-3-70B-Instruct & CoT\_FS (Q) & 8 & 4bit & 
    0.5 & 1.15 & 
    0.783 \\
    \cmidrule(lr){1-7}

    Mixtral-8x7B-v0.1 & CoT\_FS & 8 & False & 
    0.5 & 1.0 & 
    0.697 \\
    \cmidrule(lr){2-7}

    Mixtral-8x7B-v0.1 & CoT\_FS & 2 & False & 
    0.5 & 1.0 & 
    0.692 \\
    \cmidrule(lr){2-7}

    Mixtral-8x7B-v0.1 & CoT\_FS & 4 & False & 
    0.5 & 1.0 & 
    0.686 \\
    \cmidrule(lr){1-7}

    Meta-Llama-3-8B-Instruct & CoT\_FS & 4 & False & 
    0.5 & 1.0 & 
    0.672 \\
    \cmidrule(lr){2-7}

    Meta-Llama-3-8B-Instruct & CoT\_FS & 8 & False & 
    0.5 & 1.0 & 
    0.658 \\
    \cmidrule(lr){2-7}

    Meta-Llama-3-8B-Instruct & CoT\_FS & 2 & False & 
    0.5 & 1.0 & 
    0.625 \\
    \cmidrule(lr){1-7}
    
    Mistral-7B-Instruct-v0.2 & CoT\_FS & 4 & False & 
    0.5 & 1.0 & 
    0.603 \\
    \cmidrule(lr){2-7}

    Mistral-7B-Instruct-v0.2 & CoT\_FS & 8 & False & 
    0.5 & 1.0 & 
    0.597 \\
    \cmidrule(lr){2-7}
    
    Mistral-7B-Instruct-v0.2 & CoT\_FS & 2 & False & 
    0.5 & 1.0 & 
    0.594 \\
    \cmidrule(lr){1-7}

    Qwen2-7B-Instruct & CoT\_FS & 2 & False & 
    0.5 & 1.15 & 
    0.667 \\
    \cmidrule(lr){2-7}

    Qwen2-7B-Instruct & CoT\_FS & 4 & False & 
    0.5 & 1.15 & 
    0.656 \\
    \cmidrule(lr){2-7}

    Qwen2-7B-Instruct & CoT\_FS & 8 & False & 
    0.5 & 1.15 & 
    0.625 \\

    \midrule
    \multicolumn{7}{c}{\textbf{Few-shot with Random Selection}} \\
    \midrule

    Meta-Llama-3-70B-Instruct & FS Rand (Q) & 4 & 4bit & 
    0.5 & 1.15 & 
    0.800 \\
    \cmidrule(lr){2-7}

    Meta-Llama-3-70B-Instruct & FS Rand (Q) & 8 & 4bit & 
    0.5 & 1.15 & 
    0.772 \\
    \cmidrule(lr){2-7}

    Meta-Llama-3-70B-Instruct & FS Rand (Q) & 2 & 4bit & 
    0.5 & 1.15 & 
    0.769 \\
    \cmidrule(lr){1-7}

    Mixtral-8x7B-v0.1 & FS Rand & 4 & False & 
    0.5 & 1.0 & 
    0.719 \\
    \cmidrule(lr){2-7}

    Mixtral-8x7B-v0.1 & FS Rand & 8 & False & 
    0.5 & 1.0 & 
    0.711 \\
    \cmidrule(lr){2-7}

    Mixtral-8x7B-v0.1 & FS Rand & 2 & False & 
    0.5 & 1.0 & 
    0.706 \\
    \cmidrule(lr){1-7}

    Meta-Llama-3-8B-Instruct & FS Rand & 2 & False & 
    0.5 & 1.0 & 
    0.672 \\
    \cmidrule(lr){2-7}

    Meta-Llama-3-8B-Instruct & FS Rand & 8 & False & 
    0.5 & 1.0 & 
    0.672 \\
    \cmidrule(lr){2-7}

    Meta-Llama-3-8B-Instruct & FS Rand & 4 & False & 
    0.5 & 1.0 & 
    0.639 \\
    \cmidrule(lr){1-7}

    Mistral-7B-Instruct-v0.2 & FS Rand & 2 & False & 
    0.5 & 1.0 & 
    0.586 \\
    \cmidrule(lr){2-7}

    Mistral-7B-Instruct-v0.2 & FS Rand & 4 & False & 
    0.5 & 1.0 & 
    0.586 \\
    \cmidrule(lr){2-7}

    Mistral-7B-Instruct-v0.2 & FS Rand & 8 & False & 
    0.5 & 1.0 & 
    0.586 \\
    \cmidrule(lr){1-7}

    Qwen2-7B-Instruct & FS Rand & 8 & False & 
    0.5 & 1.15 & 
    0.700 \\
    \cmidrule(lr){2-7}

    Qwen2-7B-Instruct & FS Rand & 2 & False & 
    0.5 & 1.15 & 
    0.689 \\
    \cmidrule(lr){2-7}

    Qwen2-7B-Instruct & FS Rand & 4 & False & 
    0.5 & 1.15 & 
    0.683 \\

    \midrule
    \multicolumn{7}{c}{\textbf{Few-shot with Semantic Similarity}} \\
    \midrule

    Meta-Llama-3-70B-Instruct & FS Sim (Q) & 4 & 4bit & 
    0.5 & 1.15 & 
    0.817 \\
    \cmidrule(lr){2-7}

    Meta-Llama-3-70B-Instruct & FS Sim (Q) & 8 & 4bit & 
    0.5 & 1.15 & 
    0.800 \\
    \cmidrule(lr){2-7}

    Meta-Llama-3-70B-Instruct & FS Sim (Q) & 2 & 4bit & 
    0.5 & 1.15 & 
    0.792 \\
    \cmidrule(lr){1-7}

    Mixtral-8x7B-v0.1 & FS Sim & 2 & False & 
    0.5 & 1.0 & 
    0.714 \\
    \cmidrule(lr){2-7}
    
    Mixtral-8x7B-v0.1 & FS Sim & 4 & False & 
    0.5 & 1.0 & 
    0.692 \\
    \cmidrule(lr){2-7}

    Mixtral-8x7B-v0.1 & FS Sim & 8 & False & 
    0.5 & 1.0 & 
    0.675 \\
    \cmidrule(lr){1-7}

    Meta-Llama-3-8B-Instruct & FS Sim & 4 & False & 
    0.5 & 1.0 & 
    0.711 \\
    \cmidrule(lr){2-7}

    Meta-Llama-3-8B-Instruct & FS Sim & 2 & False & 
    0.5 & 1.0 & 
    0.706 \\
    \cmidrule(lr){2-7}

    Meta-Llama-3-8B-Instruct & FS Sim & 8 & False & 
    0.5 & 1.0 & 
    0.681 \\
    \cmidrule(lr){1-7}

    Mistral-7B-Instruct-v0.2 & FS Sim & 4 & False & 
    0.5 & 1.0 & 
    0.633 \\
    \cmidrule(lr){2-7}

    Mistral-7B-Instruct-v0.2 & FS Sim & 8 & False & 
    0.5 & 1.0 & 
    0.611 \\
    \cmidrule(lr){2-7}

    Mistral-7B-Instruct-v0.2 & FS Sim & 2 & False & 
    0.5 & 1.0 & 
    0.608 \\
    \cmidrule(lr){1-7}

    Qwen2-7B-Instruct & FS Sim & 8 & False & 
    0.5 & 1.15 & 
    0.731 \\
    \cmidrule(lr){2-7}

    Qwen2-7B-Instruct & FS Sim & 2 & False & 
    0.5 & 1.15 & 
    0.722 \\
    \cmidrule(lr){2-7}

    Qwen2-7B-Instruct & FS Sim & 4 & False & 
    0.5 & 1.15 & 
    0.714 \\

    \midrule
    \end{tabular}%
    }
\caption{Model Performance for \textit{RiddleSense} (Part 1). The column \textbf{Quant} indicates whether the model is quantized or not.}
\label{tab:rs_results}
\end{table*}
\begin{table*}[h!]
    \centering
    \small
    \resizebox{!}{0.5\textheight}{%
    \begin{tabular}{lc|cccc|c}
    \midrule
    \textbf{Model} & \textbf{Method} & \textbf{Num.Ex} & 
    \textbf{Quant} & 
    \textbf{Temp} &  \textbf{Rep\_Pen} &
    \textbf{Average} \\
    
    \midrule
    \multicolumn{7}{c}{\textbf{RISCORE Results}} \\
    
    \midrule
    \multicolumn{7}{c}{Llama3-70B Fewshot for QA \& Llama3-70B for distractors} \\
    \midrule
    
    Meta-Llama-3-70B-Instruct & RISCORE Sim & 2 & True & 
    0.5 & 1.15 & 
    0.792 \\
    \cmidrule(lr){2-7}

    Meta-Llama-3-70B-Instruct & RISCORE Sim & 8 & True & 
    0.5 & 1.15 & 
    0.789 \\
    \cmidrule(lr){2-7}

    Meta-Llama-3-70B-Instruct & RISCORE Sim & 4 & True & 
    0.5 & 1.15 & 
    0.783 \\
    \cmidrule(lr){1-7}

    Mixtral-8x7B-Instruct-v0.1 & RISCORE Sim & 8 & False & 
    0.5 & 1.0 & 
    0.700 \\
    \cmidrule(lr){2-7}

    Mixtral-8x7B-Instruct-v0.1 & RISCORE Sim & 4 & False & 
    0.5 & 1.0 & 
    0.689 \\
    \cmidrule(lr){2-7}

    Mixtral-8x7B-Instruct-v0.1 & RISCORE Sim & 2 & False & 
    0.5 & 1.0 & 
    0.672 \\
    \cmidrule(lr){1-7}

    Meta-Llama-3-8B-Instruct & RISCORE Sim & 4 & False & 
    0.5 & 1.0 & 
    0.722 \\
    \cmidrule(lr){2-7}

    Meta-Llama-3-8B-Instruct & RISCORE Sim & 8 & False & 
    0.5 & 1.0 & 
    0.708 \\
    \cmidrule(lr){2-7}

    Meta-Llama-3-8B-Instruct & RISCORE Sim & 2 & False & 
    0.5 & 1.0 & 
    0.692 \\    

    \cmidrule(lr){1-7}

    Mistral-7B-Instruct-v0.2 & RISCORE Sim & 2 & False & 
    0.5 & 1.0 & 
    0.600 \\
    \cmidrule(lr){2-7}

    Mistral-7B-Instruct-v0.2 & RISCORE Sim & 4 & False & 
    0.5 & 1.0 & 
    0.600 \\
    \cmidrule(lr){2-7}

    Mistral-7B-Instruct-v0.2 & RISCORE Sim & 8 & False & 
    0.5 & 1.0 & 
    0.597 \\
    \cmidrule(lr){1-7}

    Qwen2-7B-Instruct & RISCORE Sim & 8 & False & 
    0.5 & 1.0 & 
    0.731 \\
    \cmidrule(lr){2-7}

    Qwen2-7B-Instruct & RISCORE Sim & 4 & False & 
    0.5 & 1.0 & 
    0.717 \\
    \cmidrule(lr){2-7}

    Qwen2-7B-Instruct & RISCORE Sim & 2 & False & 
    0.5 & 1.0 & 
    0.697 \\
    
    \midrule
    \multicolumn{7}{c}{Llama3-70B Fewshot for QA \& Llama3-8B for distractors} \\
    \midrule

    Meta-Llama-3-70B-Instruct & RISCORE Sim & 4 & True & 
    0.5 & 1.15 & 
    0.789 \\
    \cmidrule(lr){2-7}

    Meta-Llama-3-70B-Instruct & RISCORE Sim & 2 & True & 
    0.5 & 1.15 & 
    0.786 \\
    \cmidrule(lr){2-7}

    Meta-Llama-3-70B-Instruct & RISCORE Sim & 8 & True & 
    0.5 & 1.15 & 
    0.775 \\
    \cmidrule(lr){1-7}

    Mixtral-8x7B-Instruct-v0.1 & RISCORE Sim & 2 & False & 
    0.5 & 1.0 & 
    0.719 \\
    \cmidrule(lr){2-7}

    Mixtral-8x7B-Instruct-v0.1 & RISCORE Sim & 8 & False & 
    0.5 & 1.0 & 
    0.689 \\
    \cmidrule(lr){2-7}

    Mixtral-8x7B-Instruct-v0.1 & RISCORE Sim & 4 & False & 
    0.5 & 1.0 & 
    0.686 \\
    \cmidrule(lr){1-7}

    Meta-Llama-3-8B-Instruct & RISCORE Sim & 8 & False & 
    0.5 & 1.0 & 
    0.706 \\
    \cmidrule(lr){2-7}
    
    Meta-Llama-3-8B-Instruct & RISCORE Sim & 4 & False & 
    0.5 & 1.0 & 
    0.686 \\
    \cmidrule(lr){2-7}

    Meta-Llama-3-8B-Instruct & RISCORE Sim & 2 & False & 
    0.5 & 1.0 & 
    0.681 \\
    \cmidrule(lr){1-7}

    Mistral-7B-Instruct-v0.2 & RISCORE Sim & 8 & False & 
    0.5 & 1.0 & 
    0.617 \\
    \cmidrule(lr){2-7}

    Mistral-7B-Instruct-v0.2 & RISCORE Sim & 4 & False &
    0.5 & 1.0 & 
    0.606 \\
    \cmidrule(lr){2-7}

    Mistral-7B-Instruct-v0.2 & RISCORE Sim & 2 & False & 
    0.5 & 1.0 & 
    0.603 \\
    \cmidrule(lr){1-7}

    Qwen2-7B-Instruct & RISCORE Sim & 8 & False & 
    0.5 & 1.0 & 
    0.719 \\
    \cmidrule(lr){2-7}

    Qwen2-7B-Instruct & RISCORE Sim & 4 & False & 
    0.5 & 1.0 & 
    0.697 \\
    \cmidrule(lr){2-7}

    Qwen2-7B-Instruct & RISCORE Sim & 2 & False & 0.5 & 1.0 & 0.681 \\
    \cmidrule(lr){2-7}

    \midrule
    \multicolumn{7}{c}{Llama3-8B Fewshot for QA \& Llama3-8B for distractors} \\
    \midrule

    Meta-Llama-3-70B-Instruct & RISCORE Sim & 8 & True & 
    0.5 & 1.15 & 
    0.806 \\
    \cmidrule(lr){2-7}

    Meta-Llama-3-70B-Instruct & RISCORE Sim & 2 & True & 
    0.5 & 1.15 & 
    0.792 \\
    \cmidrule(lr){2-7}

    Meta-Llama-3-70B-Instruct & RISCORE Sim & 4 & True & 
    0.5 & 1.15 & 
    0.778 \\
    \cmidrule(lr){1-7}

    Mixtral-8x7B-Instruct-v0.1 & RISCORE Sim & 4 & False & 
    0.5 & 1.0 & 
    0.714 \\
    \cmidrule(lr){2-7}

    Mixtral-8x7B-Instruct-v0.1 & RISCORE Sim & 8 & False & 
    0.5 & 1.0 & 
    0.689 \\
    \cmidrule(lr){2-7}

    Mixtral-8x7B-Instruct-v0.1 & RISCORE Sim & 2 & False & 
    0.5 & 1.0 & 
    0.681 \\
    \cmidrule(lr){1-7}

    Meta-Llama-3-8B-Instruct & RISCORE Sim & 4 & False & 
    0.5 & 1.0 & 
    0.700 \\
    \cmidrule(lr){2-7}

    Meta-Llama-3-8B-Instruct & RISCORE Sim & 2 & False & 
    0.5 & 1.0 & 
    0.689 \\
    \cmidrule(lr){2-7}

    Meta-Llama-3-8B-Instruct & RISCORE Sim & 8 & False & 
    0.5 & 1.0 & 
    0.686 \\
    \cmidrule(lr){1-7}

    Mistral-7B-Instruct-v0.2 & RISCORE Sim & 8 & False & 
    0.5 & 1.0 & 
    0.614 \\
    \cmidrule(lr){2-7}

    Mistral-7B-Instruct-v0.2 & RISCORE Sim & 4 & False & 
    0.5 & 1.0 & 
    0.600 \\
    \cmidrule(lr){2-7}

    Mistral-7B-Instruct-v0.2 & RISCORE Sim & 2 & False & 
    0.5 & 1.0 & 
    0.589 \\
    \cmidrule(lr){1-7}
    
    Qwen2-7B-Instruct & RISCORE Sim & 2 & False & 
    0.5 & 1.0 & 
    0.694 \\
    \cmidrule(lr){2-7}
    
    Qwen2-7B-Instruct & RISCORE Sim & 8 & False & 
    0.5 & 1.0 & 
    0.689 \\
    \cmidrule(lr){2-7}

    Qwen2-7B-Instruct & RISCORE Sim & 4 & False & 
    0.5 & 1.0 & 
    0.683 \\

    \midrule
    \end{tabular}%
    }
\caption{Model Performance for \textit{RiddleSense} (Part 2). The column \textbf{Quant} indicates whether the model is quantized or not.}
\label{tab:rs_results1}
\end{table*}

\end{document}